\documentclass{article}
\usepackage{nanokv_preprint,times}

\usepackage{amsmath,amsfonts,bm}

\def\eqref#1{equation~\ref{#1}}

\def\1{\bm{1}}

\DeclareMathAlphabet{\mathsfit}{\encodingdefault}{\sfdefault}{m}{sl}
\SetMathAlphabet{\mathsfit}{bold}{\encodingdefault}{\sfdefault}{bx}{n}

\usepackage{graphicx}
\usepackage{booktabs}
\usepackage{amssymb}
\usepackage{multirow}
\usepackage{array}
\usepackage{xcolor}
\usepackage{colortbl}
\usepackage{float}
\usepackage{placeins}
\usepackage{microtype}
\usepackage{hyperref}
\hypersetup{hidelinks}
\hypersetup{pdftitle={NanoKV: High-Ratio KV Cache Compression for Autoregressive Video Generation},pdfauthor={Jiacheng Lu, Haoyi Zhu, Sipei Yi, Enze Xie, Cheng Zhuo, Yu Li}}
\definecolor{codeblue}{HTML}{063BEE}
\DeclareRobustCommand{\iconhref}[3]{%
  \texorpdfstring{\href{#1}{\raisebox{-0.15em}{\includegraphics[height=1em]{#2}}\,\textcolor{codeblue}{#3}}}{#3}%
}
\usepackage{url}

\definecolor{OursRowBlue}{RGB}{229,240,253}
\definecolor{OursComputeBlue}{RGB}{216,232,251}

\title{NanoKV: High-Ratio KV Cache Compression\\for Autoregressive Video Generation}

\author{Jiacheng Lu$^{1}$ \quad Haoyi Zhu$^{2}$ \quad Sipei Yi$^{1}$ \quad
Enze Xie$^{2}$ \quad Cheng Zhuo$^{1}$ \quad Yu Li$^{1}$\thanks{Corresponding author: Yu Li at \href{mailto:yu.li.sallylee@gmail.com}{yu.li.sallylee@gmail.com}.} \\
$^{1}$Zhejiang University \quad $^{2}$NVIDIA}

\begin{document}
\maketitle
\centerline{%
\iconhref{https://jiacheng-lu03.github.io/NanoKV/}{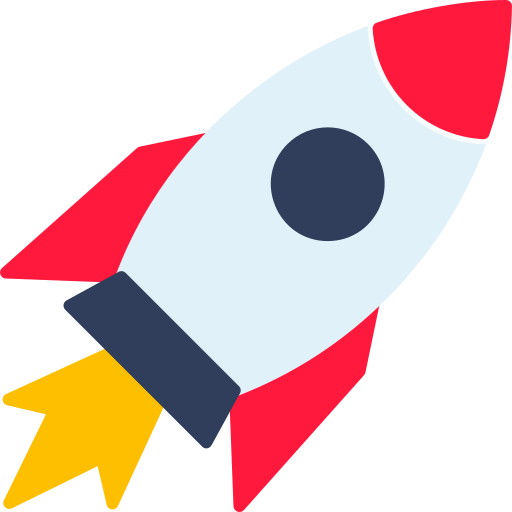}{Project Page}
\quad
\iconhref{https://github.com/jiacheng-lu03/NanoKV}{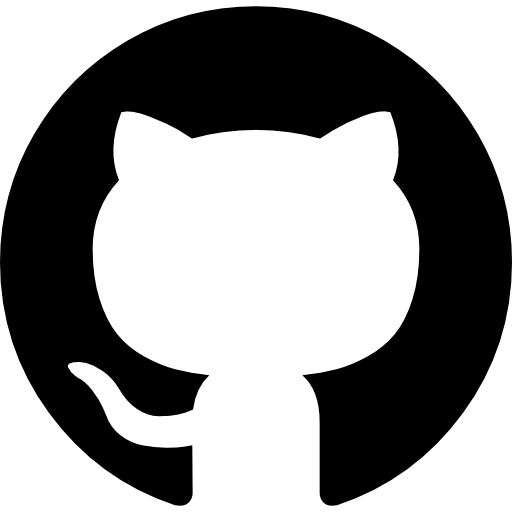}{GitHub}
}

\begingroup
\setlength{\intextsep}{6pt}
\setlength{\abovecaptionskip}{6pt}
\begin{figure}[H]
\centering
\includegraphics[width=\textwidth]{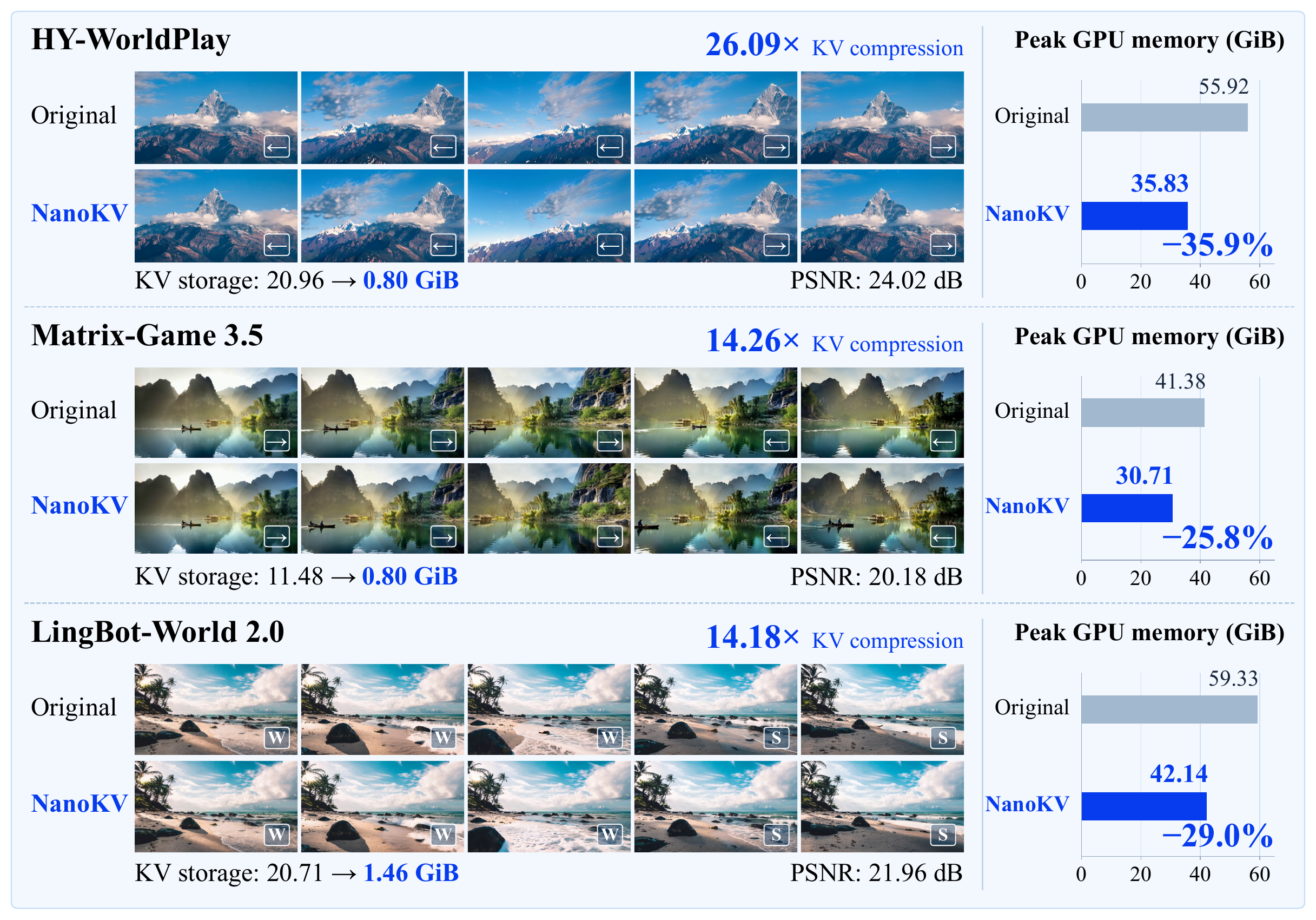}
\caption{NanoKV achieves high-ratio historical KV compression while preserving generated video quality across three autoregressive video backbones. Displayed PSNR is measured against the matched Original rollout. NanoKV reaches up to 26.09$\times$ effective KV compression and reduces measured peak GPU memory by up to 35.9\%.}
\label{fig:teaser}
\end{figure}
\endgroup

\begin{abstract}
Autoregressive video models rely on historical key--value (KV) caches to preserve scene consistency over extended rollouts, making KV storage and repeated attention over history a growing memory and computational bottleneck. Aggressive compression can degrade generation quality, while reconstructing full-dimensional KV leaves attention computation largely unchanged. We introduce NanoKV, a frozen-backbone framework for generation-aware high-ratio KV compression. NanoKV exploits redundancy across tokens and attention heads by storing shared content in cluster centers and encoding joint head--channel residuals into low-bit coefficients. Crucially, NanoKV uses different fitting objectives for Keys and Values according to their distinct roles in attention: Key decoders are optimized to preserve downstream block outputs, whereas Value encoders and decoders are fitted for faithful reconstruction through the deployed quantization path. Building on the same compressed representation, the optional NanoKV-Fast variant reuses the stored coefficients for mixed-dimensional attention, reducing computation without modifying the video backbone. Across three autoregressive video backbones, NanoKV consistently improves the quality--storage trade-off. On HY-WorldPlay, NanoKV achieves $26.09\times$ effective KV compression---nearly twice that of KVTC---while improving PSNR by 1.0 dB. NanoKV-Fast retains $20.53\times$ effective compression and achieves an approximately $1.15\times$ end-to-end speedup over KVTC.

\end{abstract}

\section{Introduction}
\label{sec:introduction}


Autoregressive video models rely on historical Key and Value (KV) caches to preserve objects, layout, and appearance over extended rollouts and when revisiting earlier views~\citep{worldplay2025,matrixgame35,lingbotworld2}. 
Growing history creates a systems bottleneck: retaining KV consumes GPU memory, while repeatedly attending to it increases computation.
Reducing the history window or evicting cached tokens lowers these costs, but also removes scene information that may become important later in the rollout.


KV feature compression offers a complementary solution: quantization and low-rank projection reduce each token's bitwidth or feature dimension without shortening history~\citep{qvg2026,kvtc2026}. 
\textbf{At high compression ratios, however, two challenges become critical.
}First, \emph{compression must preserve not only the reconstructed KV features, but also their downstream effect on generation}. KV reconstruction errors perturb attention outputs and propagate through subsequent Transformer blocks, so minimizing KV reconstruction error alone does not fully capture compression-induced generation distortion.
Second, \emph{compact storage does not automatically yield compact computation}. 
Compressed storage saves memory, but full-dimensional reconstruction leaves historical-attention computation largely unchanged.

We introduce \textbf{\textit{NanoKV}}, a frozen-backbone framework for high-ratio historical KV compression. 
Our key observation is that effective compression should preserve how historical KV is \emph{used} by the generator, not merely how accurately the cache tensors are reconstructed. 
NanoKV first exploits redundancy in video KV by jointly encoding token-specific residuals relative to shared centers across heads and channels, yielding a compact low-bit representation without discarding historical tokens. 
To preserve generation under aggressive compression, we further adapt the Key and Value encoders and decoders (codecs) according to their distinct roles in attention: Key decoders are optimized through block-output alignment to preserve downstream Transformer behavior, while Value codecs are fitted through the deployed quantization path for faithful low-bit reconstruction. 
Together, these enable substantially higher compression without retraining the backbone.

The same compressed representation also provides a direct path to lower computation. 
Instead of reconstructing all history at full dimension, optional \textbf{\textit{NanoKV-Fast}} reuses the coefficients for mixed-dimensional attention, assigning lower dimensions to less sensitive history and higher or full dimensions where needed. 
The same compressed history thus supports both high-fidelity reconstruction and compute-efficient attention.


\begin{figure}[!t]
\centering
\includegraphics[width=0.94\linewidth]{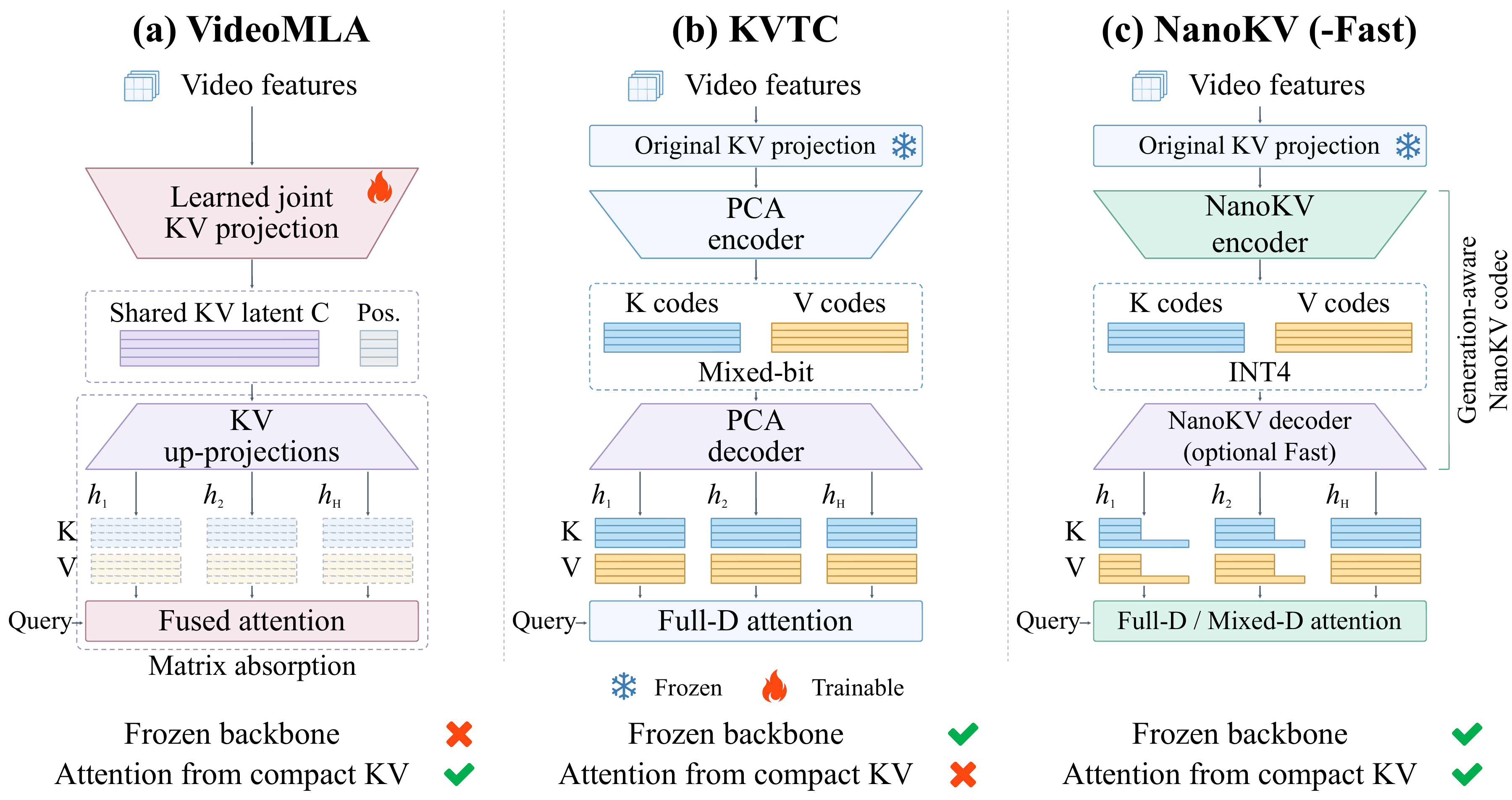}
\caption{\textbf{Comparison of NanoKV(-Fast) with representative KV compression and attention-access baselines.} (a) VideoMLA modifies KV projections and adapts the model for shared latent-space attention. (b) KVTC preserves the original projections but reconstructs full-dimensional KV before attention. (c) NanoKV preserves the original projections and frozen backbone, using generation-aware, role-specific K/V fitting; optional NanoKV-Fast reuses the coefficients for mixed-dimensional attention. $h_1,\ldots,h_H$ denote attention heads.}
\label{fig:compression-designs}
\end{figure}

Our main contributions are:
\begin{itemize}
    \item \textbf{Generation-aware high-ratio KV compression.}
    We formulate high-ratio historical KV compression around preserving the downstream effect of compressed history on generation, rather than KV reconstruction fidelity alone.
    This perspective motivates compression objectives that account for how the frozen video backbone uses historical KV.

    \item \textbf{Structure- and role-aware KV compression.}
    We instantiate this principle in NanoKV through shared-center residual coding, joint head--channel compression, and role-specific K/V fitting.
    Key decoders are optimized to preserve downstream block outputs, while Value codecs are fitted through the deployed quantization path for faithful low-bit reconstruction.

    \item \textbf{Turning compressed storage into efficient attention.}
    The optional NanoKV-Fast variant directly constructs mixed-dimensional attention features from NanoKV's compressed coefficients, reducing historical-attention computation without modifying the backbone.


\end{itemize}

Across three autoregressive video backbones, NanoKV consistently improves the quality--storage trade-off, reaching up to $26.09\times$ effective KV compression. NanoKV-Fast achieves lower end-to-end latency than all evaluated competing compression baselines.

\section{Related Work}
\label{sec:related}

\textbf{Selecting historical tokens.}
A common way to reduce KV cost is to retain only a subset of the history.
Recent methods exploit temporal distance, future relevance, frame content, and head specialization to identify useful visual history~\citep{packcache2026,pafukv2026,forcingkv2026,focusedforcing2026,futureforcing2026,dummyforcing2026}.
TempCache in FAST-AR groups temporally corresponding KV and retains recent representatives, WorldKV retrieves scene-relevant history, and Sparse Forcing learns persistent sparse memory~\citep{tempcache2026,worldkv2026,sparseforcing2026}.
Unlike these methods that select \emph{which} tokens to retain, NanoKV preserves the history policy and compresses retained KV features.

\textbf{Compressing retained KV features.}
KV feature compression reduces the storage cost of each retained token through lower precision or lower-dimensional representations.
Quantization methods control low-bit distortion using Key/Value statistics, activation outliers, or attention-related criteria~\citep{kivi2024,kvquant2024,wkvquant2024,turboquant2025}, while low-rank methods project KV features into compact spaces using fixed or learned mappings~\citep{palu2024,eigenattention2024,matryoshkakv2025}.
For autoregressive video generation, QVG exploits shared visual centers and residual quantization~\citep{qvg2026}.
Originally developed for LLM inference, KVTC combines PCA-based feature compression with adaptive mixed-bit quantization; we adapt it to video models as our primary feature-compression baseline~\citep{kvtc2026}.
At high compression ratios, reconstruction fidelity alone does not fully characterize downstream generation. NanoKV therefore combines joint head--channel compression with generation-aware, role-specific K/V fitting.


\textbf{Computing from compact KV.}
Compact KV storage does not automatically reduce attention computation when compressed features are reconstructed to full dimension before use.
In autoregressive video generation, VideoMLA modifies the KV projections and adapts the model for latent-space attention~\citep{videomla2026}.
Separately, sparse video attention reduces computation by limiting Query--Key interactions~\citep{svg22025}.
Unlike VideoMLA's model adaptation, optional NanoKV-Fast constructs mixed-dimensional KV from NanoKV's compressed coefficients while keeping the backbone frozen (Figure~\ref{fig:compression-designs}).
\section{Motivation}
\label{sec:motivation}

High-ratio video KV compression raises three questions: where redundancy should be removed, what fidelity the compressed representation should preserve, and whether the same compact state can also reduce attention computation.

\begin{figure*}[t]
  \centering
  \includegraphics[width=0.9\textwidth]{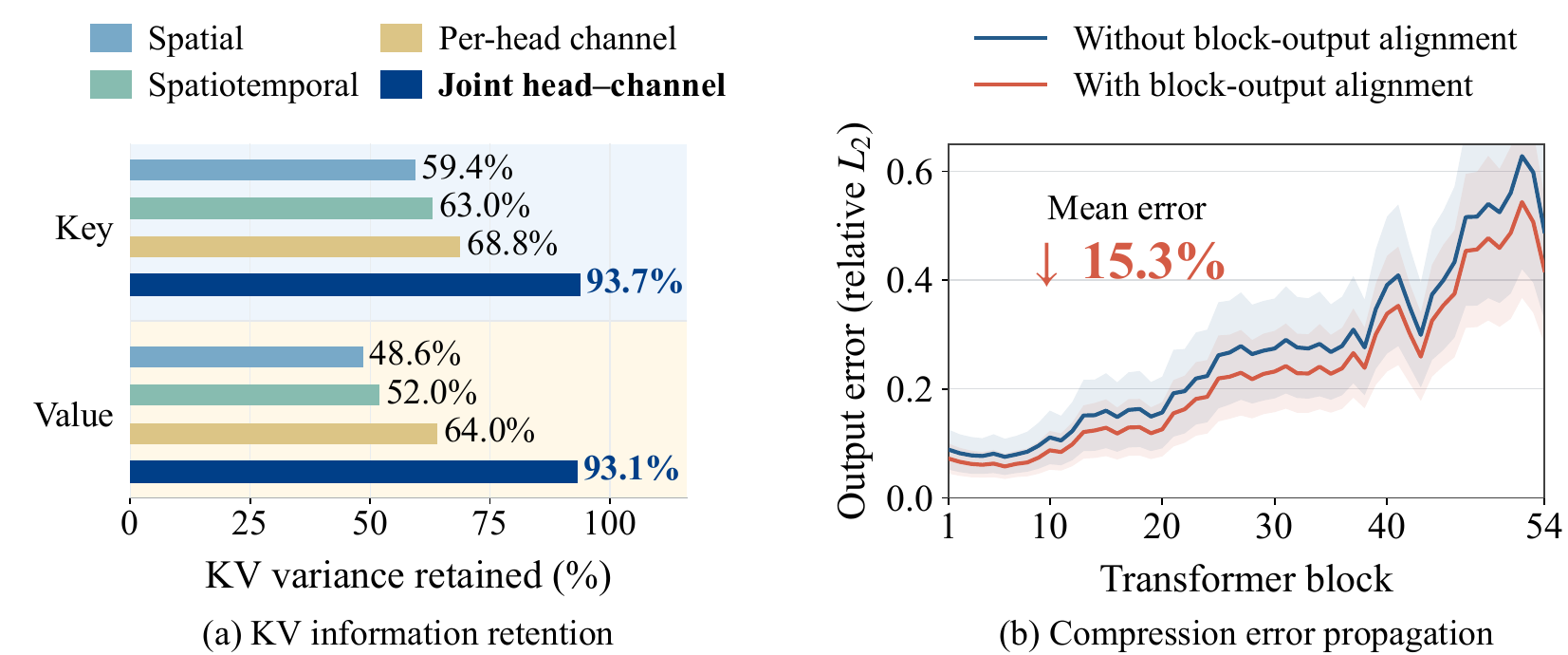}
  \caption{\textbf{Motivation for generation-aware KV compression.}
(a) Joint head--channel PCA retains the most KV variance at matched $8\times$ dimensional compression.
(b) Compression-induced output errors accumulate with network depth, while block-output alignment reduces mean relative error by 15.3\%. Errors are measured relative to Original; lines and shading show the mean and standard deviation across 16 chunks. Results are on HY-WorldPlay.}
  \label{fig:motivation}
\end{figure*}



\noindent\textbf{Where to compress: exploiting cross-head redundancy.}
Video KV exhibits strong spatial and temporal redundancy~\citep{qvg2026}, while prior work on LLM KV compression suggests that attention heads also contain correlated feature structure~\citep{kvtc2026}.
Whether this cross-head redundancy effectively transfers to video KV remains unclear.
We therefore compare spatial, spatiotemporal, per-head channel, and joint head--channel PCA under the same 8$\times$ dimensional budget.
Figure~\ref{fig:motivation}(a) shows that joint head--channel compression retains 93.7\% of Key variance and 93.1\% of Value variance, substantially more than the alternatives.
This supports joint cross-head compression, but retained variance alone does not establish fidelity to downstream generation.



\noindent\textbf{What to preserve: downstream generation behavior.}
Compression errors perturb attention outputs; the resulting discrepancies propagate through subsequent Transformer blocks and interact with new errors.
For Keys, let $\delta_\ell$ denote the feature difference entering block $\ell$ between paths using original and compressed history.
A first-order approximation gives
\begin{equation}
    \delta_{\ell+1} \approx J_\ell \delta_\ell + \epsilon_\ell,
    \label{eq:rollout-unroll}
\end{equation}
Here the local block Jacobian $J_\ell$ propagates existing discrepancies, while $\epsilon_\ell$ is the new output error from compressed history at a fixed block input (Appendix~\ref{app:layer-error-propagation}).
Thus, the downstream effect of compression depends not only on KV reconstruction fidelity, but also on how reconstruction errors are transformed and accumulated through the network.
We examine whether fitting the Key decoder to match block outputs under original and compressed history can reduce these discrepancies (Section~\ref{sec:key-learning}).
Figure~\ref{fig:motivation}(b) illustrates this effect: output discrepancies generally grow with network depth, while block-output alignment reduces the mean error by 15.3\%.
These observations motivate \emph{generation-aware} codec fitting beyond KV reconstruction alone.
Because Keys and Values play different roles in attention, we further use role-specific objectives for their codecs.



\noindent\textbf{How to compute: from compact storage to compact attention.}
Compact storage alone does not reduce attention computation if historical KV is reconstructed at full dimension. The compressed representation enables lower-dimensional attention, but different heads and historical tokens tolerate dimensionality reduction differently. This motivates NanoKV-Fast to use mixed-dimensional attention, preserving higher dimensions only for sensitive parts.
\section{Method}
\label{sec:method}

\begin{figure*}[!t]
  \centering
  \includegraphics[width=\textwidth]{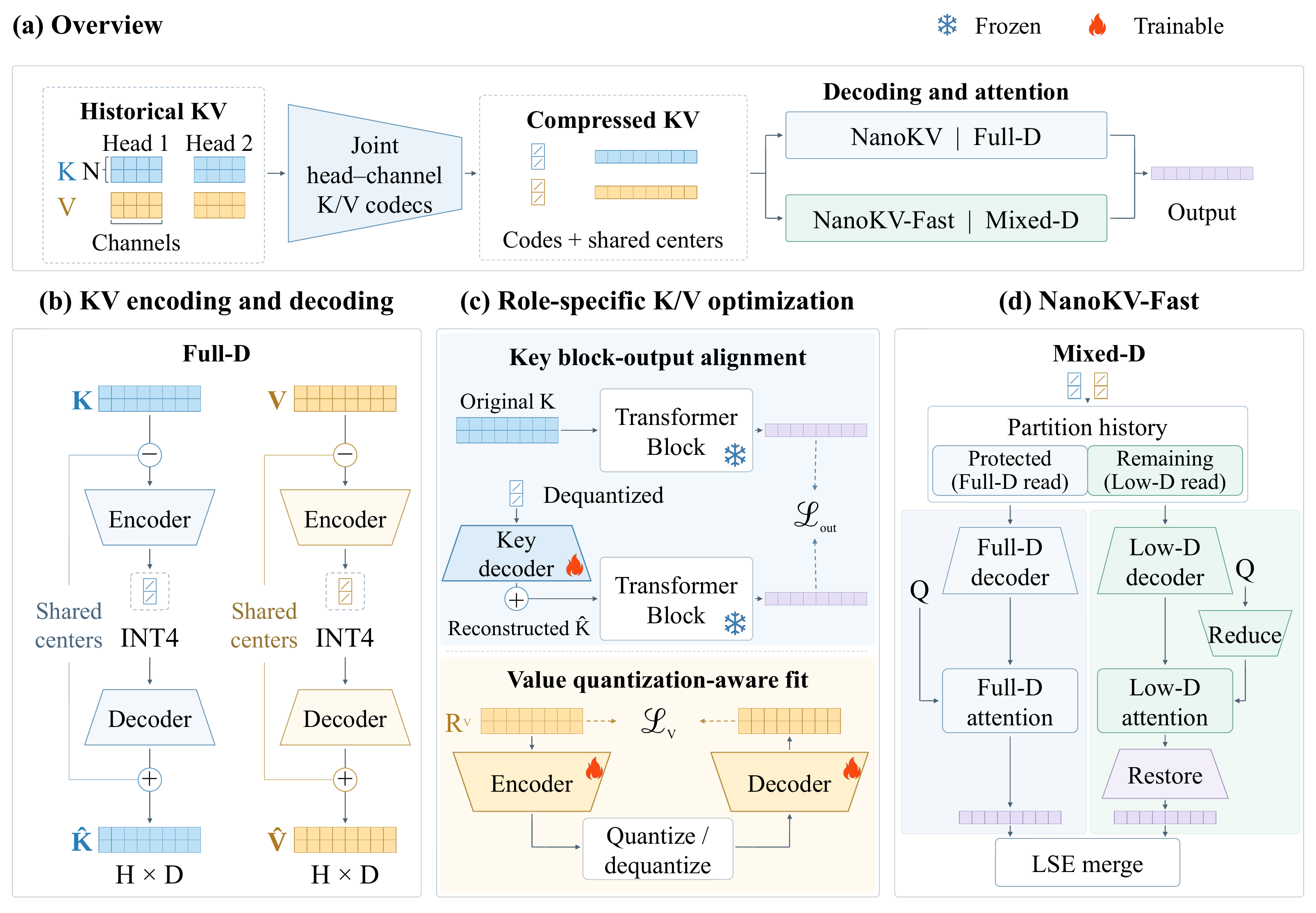}
  \caption{Overview of NanoKV for generation-aware high-ratio KV compression. (a) Joint head--channel encoding stores historical KV as coefficients and shared centers, read through full-dimensional (NanoKV) or mixed-dimensional (NanoKV-Fast) decoding and attention. (b) The encoder compresses residuals into INT4 coefficients; the decoder reconstructs them and adds back centers. (c) Key decoders use block-output alignment; Value encoders and decoders use quantization-aware fitting. (d) A fixed policy routes history to full- or low-dimensional attention. The latter reduces Queries and restores output dimensions before LSE merging.}
  \label{fig:method}
\end{figure*}

NanoKV combines joint head--channel compression (Section~\ref{sec:representation}) with generation-aware, role-specific K/V optimization (Section~\ref{sec:role-aware}), keeping the backbone frozen. Optional NanoKV-Fast reuses this representation for mixed-dimensional attention (Section~\ref{sec:latent-attention}); Figure~\ref{fig:method} outlines the framework.

\subsection{Joint head--channel KV compression}
\label{sec:representation}
Motivated by the cross-head redundancy in Section~\ref{sec:motivation}, we concatenate each token's $H$ heads with $D$ channels into $x_i\in\mathbb R^{HD}$ and compress them jointly, without removing tokens (Figure~\ref{fig:method}(b)). Keys and Values share this representation structure but use separate, layer-specific encoder--decoder pairs (codecs), fitted with role-specific objectives (Section~\ref{sec:role-aware}). We omit layer indices and K/V superscripts below.

\paragraph{Shared-center residual encoding.}
Adapting QVG's center-based residual representation~\citep{qvg2026} to the joint head--channel space, we cluster each incoming historical segment using online K-means. Each token stores its center index $a_i$ and encodes $x_i-c_{a_i}$; centers are shared within the segment. Thus, coefficients represent token differences.

\paragraph{Encoding and decoding.}
The encoder $\mathbf E\in\mathbb R^{HD\times r}$ maps residuals to $r$ coefficients, further compressed with INT4 quantization. The decoder $W_{\mathrm{dec}}\in\mathbb R^{r\times HD}$ reconstructs features:
\begin{equation}
 \begin{aligned}
 \text{Encode:}\quad &z_i=(x_i-c_{a_i}-\mu)\mathbf E,\qquad q_i=\mathcal Q_4(z_i;s_i),\\
 \text{Decode:}\quad &\hat z_i=s_iq_i,\qquad \hat x_i=c_{a_i}+\mu+\hat z_iW_{\mathrm{dec}} .
 \end{aligned}
 \label{eq:codec-path}
\end{equation}
Here $\mu$ is the fixed residual mean, $s_i$ is a token-wise scale, and $\mathcal Q_4$ denotes symmetric INT4 quantization. The cache stores packed codes, scales, center indices, and centers. NanoKV reconstructs full-dimensional KV for attention.

\subsection{Role-specific optimization for Keys and Values}
\label{sec:role-aware}
A compact representation alone does not ensure faithful downstream generation. Keys determine attention weights, so their reconstruction errors can change how historical information is selected and combined. We therefore fit Key decoders through block-output alignment to constrain downstream discrepancies beyond local KV reconstruction error (Section~\ref{sec:motivation}). Values provide the content aggregated by attention; for given attention weights, their reconstruction errors directly affect the aggregated output. We therefore fit Value encoders and decoders through the deployed quantization path to preserve this content (Figure~\ref{fig:method}(c)).

\paragraph{Block-output alignment for Key decoders.}
\label{sec:key-learning}
After PCA initialization, we fix the encoder and coefficients and train the Key decoder through layer-by-layer block-output alignment.

We train consecutive blocks in groups. Two paths share the group input but use original or reconstructed historical Keys, respectively, with outputs aligned at each block. Their discrepancies reflect both local compression errors and errors propagated within the group. For their layer-$\ell$ outputs $Y_\ell$ and $\hat Y_\ell$, we minimize
\begin{equation}
 \mathcal L_K=\mathcal L_{\mathrm{out}}(\hat Y_\ell,Y_\ell)
 +\mathcal L_{K,\mathrm{rec}}(\hat K_\ell,K_\ell).
 \label{eq:teacher-student}
\end{equation}
The terms constrain block-output discrepancy and Key reconstruction. Groups are trained from shallow to deep, using compressed-history outputs from completed groups as subsequent inputs. Only Key decoders are updated; the backbone stays frozen.

\paragraph{Quantization-aware fitting for Value encoders and decoders.}
\label{sec:value-learning}
Quantizing only after fitting would introduce distortion absent from training. We instead include INT4 rounding in the forward pass and jointly train the Value encoder and decoder to reconstruct the center-subtracted residual $R^V$:
\begin{equation}
 \mathcal L_V=\mathcal L_{\mathrm{rec}}^V(\hat R^V,R^V),
 \label{eq:value-qat-main}
\end{equation}
where $\hat R^V$ is the reconstructed residual before adding the center in Equation~\ref{eq:codec-path}. Both maps use random orthogonal initialization and an identity approximation to the rounding gradient; the backbone and Key branch stay fixed. Appendix~\ref{app:codec-objectives} details the losses and training.

\subsection{NanoKV-Fast: attention from compressed features}
\label{sec:latent-attention}
To also reduce attention computation (Section~\ref{sec:motivation}), optional NanoKV-Fast reuses the compressed coefficients for mixed-dimensional KV, retaining higher or full dimensions for sensitive parts.

\paragraph{Low-dimensional attention.}
The joint decoder has head-specific output blocks, from which we construct low-dimensional decoders for selected heads. Let $Z_K,Z_V$ contain dequantized coefficients for selected history. For a head using $d_h<D$ dimensions, decoders $B_h^K,B_h^V\in\mathbb R^{r\times d_h}$ construct KV features and $A_h\in\mathbb R^{D\times d_h}$ reduces Queries:
\[
 \bar Q_h=Q_hA_h,\qquad
 \bar K_h=Z_KB_h^K+\bar C_h^K,\qquad
 \bar V_h=Z_VB_h^V+\bar C_h^V.
\]
The offsets map the stored centers and means to the same low-dimensional spaces. Attention aggregates Values before $L_h\in\mathbb R^{d_h\times D}$ restores the output dimension:
\begin{equation}
 \bar O_h=\operatorname{softmax}\!\left(
 \alpha_h\bar Q_h\bar K_h^\top/\sqrt D\right)\bar V_h,
 \qquad O_h^{\mathrm{low}}=\bar O_hL_h .
 \label{eq:latent-attention}
\end{equation}
We retain $1/\sqrt D$ because low-dimensional scores approximate the original scores; $\alpha_h$ corrects scale mismatch. Attention arithmetic for this history is approximately $d_h/D$ of the original. The maps are fitted against full-dimensional attention on the same compressed cache, with the codec and backbone fixed; model-specific output-mean compensation is detailed in Appendix~\ref{app:latent-attention-details}.

\paragraph{Protecting sensitive history.}
Sensitivity analysis and rollout-quality feedback determine a fixed protection policy per model, routing sensitive layers, heads, and historical tokens to higher or full dimensions and the remainder to lower dimensions (Appendix~\ref{app:learning-protocol}, Table~\ref{tab:protection-configuration}). Log-sum-exp (LSE) statistics combine low- and full-dimensional branch outputs under one softmax normalization (Figure~\ref{fig:method}(d)). Fitting and implementation details are in Appendices~\ref{app:learning-protocol}--\ref{app:latent-attention-details}.

\section{Experiments}
\label{sec:experiments}

\begin{table}[!t]
\caption{Quality, compression, and efficiency comparison of NanoKV and its Fast variant with baseline methods. Parentheses denote nominal coefficient compression ratios.}
\label{tab:main-results}
\centering
\setlength{\tabcolsep}{6pt}
\renewcommand{\arraystretch}{1.12}
\footnotesize
\resizebox{0.98\textwidth}{!}{%
\begin{tabular}{lrrrrrrr}
\toprule
\begin{tabular}[c]{@{}l@{}}\textbf{Method}\end{tabular} & \begin{tabular}[c]{@{}c@{}}\textbf{PSNR}$\uparrow$\end{tabular} & \begin{tabular}[c]{@{}c@{}}\textbf{SSIM}$\uparrow$\end{tabular} & \begin{tabular}[c]{@{}c@{}}\textbf{LPIPS}$\downarrow$\end{tabular} & \begin{tabular}[c]{@{}c@{}}\textbf{VBench}$\uparrow$\end{tabular} & \begin{tabular}[c]{@{}c@{}}\textbf{Total KV}$\downarrow$\\(GiB)\end{tabular} & \begin{tabular}[c]{@{}c@{}}\textbf{KV CR}$\uparrow$\end{tabular} & \begin{tabular}[c]{@{}c@{}}\textbf{E2E Time /}\\\textbf{Speedup}$\uparrow$\end{tabular} \\
\midrule
\textit{HY-WorldPlay} & -- & -- & -- & 0.8129 & 20.96 & 1.00$\times$ & 437.63\,s \\
\cmidrule(lr){1-8}
Quant VideoGen & 17.58 & 0.5379 & 0.3887 & 0.7744 & 3.14 & 6.69$\times$ & 0.86$\times$ \\
TurboQuant & 17.95 & 0.5459 & 0.3523 & 0.7918 & 4.18 & 5.01$\times$ & 0.90$\times$ \\
KVTC (32$\times$) & 18.56 & 0.5796 & 0.2900 & 0.7936 & 1.55 & 13.50$\times$ & 0.92$\times$ \\
\rowcolor[HTML]{DDF1FF}[\dimexpr\tabcolsep+0.2pt\relax][\dimexpr\tabcolsep+0.2pt\relax] \textbf{NanoKV} (32$\times$) & \textbf{19.56} & \textbf{0.6152} & \textbf{0.2399} & \textbf{0.8057} & \textbf{0.80} & \textbf{26.09$\times$} & 0.96$\times$ \\
\rowcolor[HTML]{C5E5FC}[\dimexpr\tabcolsep+0.2pt\relax][\dimexpr\tabcolsep+0.2pt\relax] \textbf{NanoKV-Fast} (32$\times$) & 19.04 & 0.5931 & 0.2787 & 0.7981 & 1.02 & 20.53$\times$ & \textbf{1.06}$\times$ \\
\midrule
\textit{Matrix-Game 3.5} & -- & -- & -- & 0.8293 & 11.48 & 1.00$\times$ & 306.70\,s \\
\cmidrule(lr){1-8}
Quant VideoGen & 16.52 & 0.5150 & 0.4115 & 0.7782 & 2.89 & 3.97$\times$ & 0.91$\times$ \\
TurboQuant & 16.73 & 0.5314 & 0.3867 & 0.7951 & 2.29 & 5.01$\times$ & 0.91$\times$ \\
KVTC (32$\times$) & 18.83 & 0.6347 & \textbf{0.2380} & \textbf{0.8302} & 1.24 & 9.28$\times$ & 0.57$\times$ \\
\rowcolor[HTML]{DDF1FF}[\dimexpr\tabcolsep+0.2pt\relax][\dimexpr\tabcolsep+0.2pt\relax] \textbf{NanoKV} (32$\times$) & \textbf{19.08} & \textbf{0.6377} & 0.2425 & 0.8273 & \textbf{0.80} & \textbf{14.26$\times$} & 0.92$\times$ \\
\rowcolor[HTML]{C5E5FC}[\dimexpr\tabcolsep+0.2pt\relax][\dimexpr\tabcolsep+0.2pt\relax] \textbf{NanoKV-Fast} (32$\times$) & 18.79 & 0.6280 & 0.2498 & 0.8280 & 0.96 & 11.94$\times$ & \textbf{1.04}$\times$ \\
\midrule
\textit{LingBot-World 2.0} & -- & -- & -- & 0.8254 & 20.71 & 1.00$\times$ & 132.40\,s \\
\cmidrule(lr){1-8}
Quant VideoGen & 20.46 & 0.6193 & 0.2310 & \textbf{0.8256} & 4.30 & 4.81$\times$ & 0.87$\times$ \\
TurboQuant & 19.06 & 0.5650 & 0.2890 & 0.8235 & 4.59 & 4.51$\times$ & 0.87$\times$ \\
KVTC (32$\times$) & 20.10 & 0.6150 & 0.2431 & 0.8228 & 4.73 & 4.38$\times$ & 0.79$\times$ \\
\rowcolor[HTML]{DDF1FF}[\dimexpr\tabcolsep+0.2pt\relax][\dimexpr\tabcolsep+0.2pt\relax] \textbf{NanoKV} (32$\times$) & \textbf{20.81} & \textbf{0.6352} & \textbf{0.2156} & 0.8252 & \textbf{1.46} & \textbf{14.18$\times$} & 0.93$\times$ \\
\rowcolor[HTML]{C5E5FC}[\dimexpr\tabcolsep+0.2pt\relax][\dimexpr\tabcolsep+0.2pt\relax] \textbf{NanoKV-Fast} (32$\times$) & 20.68 & 0.6314 & 0.2183 & 0.8249 & 1.78 & 11.62$\times$ & \textbf{0.98}$\times$ \\
\bottomrule
\end{tabular}
}
\end{table}

\subsection{Setup}
\paragraph{Models.}
To assess generation-aware high-ratio KV compression across backbones, we evaluate NanoKV and its optional Fast variant on three autoregressive video models: HY-WorldPlay~\citep{worldplay2025}, Matrix-Game 3.5~\citep{matrixgame35}, and LingBot-World 2.0~\citep{lingbotworld2}. Within each model, all methods use the same inputs, camera actions, resolution, and cache history to isolate the effect of KV compression and attention access.

\paragraph{Baselines.}
We compare with the quantization methods QVG~\citep{qvg2026} and TurboQuant~\citep{turboquant2025}, and the low-rank compression method KVTC~\citep{kvtc2026}. The uncompressed model (Original) provides the reference rollout and runtime baseline. The same cache-lifecycle optimizations are applied across methods (Appendix~\ref{app:cache-protocol}).

\paragraph{Datasets.}
The NanoKV and KVTC compression representations are fitted on 48 trajectories from 24 fixed images with two actions each; QVG and TurboQuant require no prior data fitting. Evaluation uses 100 VBench-I2V images with four camera actions each, averaging quality over the resulting 400 videos per configuration. Appendix~\ref{app:learning-protocol} details NanoKV-Fast map fitting and configuration selection, using development images disjoint from the evaluation set.

\paragraph{Metrics.}
PSNR, SSIM, and LPIPS measure compression-induced deviations from matched Original rollouts, rather than from ground-truth videos. To assess video quality beyond reference fidelity, we also report VBench~\citep{vbench2023}, averaging subject consistency, background consistency, motion smoothness, aesthetic quality, and imaging quality. KV cache size and compression ratio include each method's additional storage. We report Original's end-to-end latency in seconds and each method's speedup as Original latency divided by method latency. Peak GPU memory is reported in Appendix~\ref{app:runtime-memory}.

\paragraph{Implementation.}
Experiments run on a single NVIDIA A100 GPU, with the video backbone frozen throughout codec fitting and low-dimensional attention adaptation. Fitting settings and NanoKV-Fast implementation details are provided in Appendices~\ref{app:codec-objectives} and~\ref{app:learning-protocol}.

\subsection{Comparison with Baselines}

Table~\ref{tab:main-results} compares quality, storage, and latency. At nominal $32\times$ on HY-WorldPlay, NanoKV nearly doubles KVTC's effective compression with about 1.0 dB higher PSNR. On LingBot-World 2.0, it also improves reference fidelity while using substantially less KV memory.

On Matrix-Game 3.5, NanoKV and KVTC have comparable reference fidelity, but NanoKV achieves greater effective compression and lower runtime overhead. KVTC's larger reconstruction matrices add storage and computation costs (Appendix~\ref{app:reconstruction-cost}).

NanoKV-Fast reduces compression overhead, exceeding Original's end-to-end speed on HY-WorldPlay and Matrix-Game 3.5 but remaining slightly slower on LingBot-World 2.0. Gains depend on low-dimensional budgets, historical attention's runtime share, and decoding and mapping overhead. Appendix Figure~\ref{fig:system-storage} separates cached data and resident operators; Appendix~\ref{app:video-comparisons} provides visual comparisons.

\subsection{Understanding the Quality--Compression--Efficiency Trade-off}
\label{sec:analysis}

\paragraph{What preserves fidelity under high-ratio compression?}
Table~\ref{tab:component-ablation} adds shared-center residual coding, Value fitting, and Key alignment to joint PCA with INT4, using 20 HY-WorldPlay videos from Table~\ref{tab:main-results}.
\begin{table}[!t]
\caption{Component analysis on HY-WorldPlay at nominal $16\times$ compression.}
\label{tab:component-ablation}
\centering
\footnotesize
\setlength{\tabcolsep}{6pt}
\renewcommand{\arraystretch}{1.12}
\begin{tabular}{lrrrr}
\toprule
\textbf{Configuration} & \textbf{PSNR}$\uparrow$ & \textbf{SSIM}$\uparrow$ & \textbf{LPIPS}$\downarrow$ & \textbf{VBench}$\uparrow$ \\
\midrule
Joint PCA + INT4 & 16.38 & 0.4953 & 0.4077 & 0.7753 \\
$+$ Shared centers & 18.38 & 0.5754 & 0.3047 & 0.7949 \\
$+$ Value quantization-aware fitting & 18.54 & 0.5839 & 0.2922 & 0.8007 \\
\rowcolor[HTML]{DDF1FF}[\dimexpr\tabcolsep+0.2pt\relax][\dimexpr\tabcolsep+0.2pt\relax]
$+$ \textbf{Key block-output alignment (full)} & \textbf{20.20} & \textbf{0.6692} & \textbf{0.1968} & \textbf{0.8171} \\
\bottomrule
\end{tabular}
\end{table}

Shared centers improve fidelity by separating shared content from token-specific residuals.
Fitting Values through deployed quantization further improves fidelity by targeting content reconstruction.
Key alignment adds gains from downstream output constraints beyond reconstruction, consistent with Section~\ref{sec:motivation}.
Together, shared-center residual coding and role-specific K/V fitting help preserve generation fidelity under high-ratio compression.

\begin{figure}[!t]
 \centering
 \includegraphics[width=\textwidth,trim=0 26bp 0 29bp,clip]{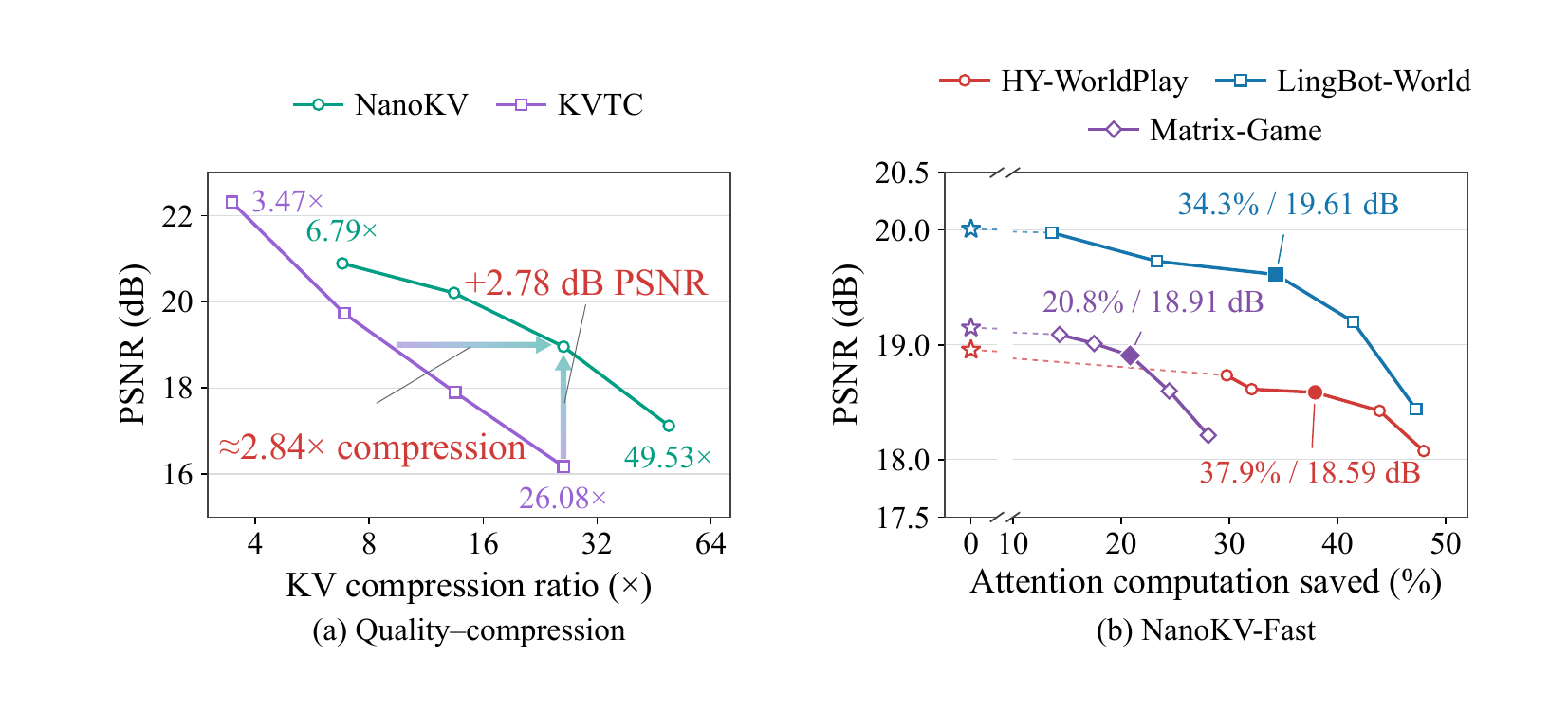}
 \caption{\textbf{Balancing fidelity, compression, and computation.} (a) NanoKV improves the HY-WorldPlay quality--compression frontier over KVTC (videos from Table~\ref{tab:component-ablation}); arrows compare matched quality and near-matched compression. (b) NanoKV-Fast trades fidelity for historical-attention computation savings, not end-to-end time. Stars denote NanoKV; filled markers indicate main-table Fast configurations. Each model uses 20 fixed videos.}
 \label{fig:compute-tradeoff}
\end{figure}

\paragraph{How do compression and computation trade off with fidelity?}
We next examine fidelity across compression and computation settings.
Figure~\ref{fig:compute-tradeoff}(a) shows NanoKV's better quality--compression frontier: 2.78 dB higher PSNR at approximately $26.08\times$ effective compression, and an estimated $2.84\times$ compression-ratio advantage at 19 dB (linear interpolation in log-compression space).
Panel~(b) presents the quality--efficiency trade-off within each backbone by varying the history routed to low-dimensional attention while keeping the fitted maps fixed.
More aggressive routing saves more historical-attention computation but lowers generation fidelity; the main table reports moderate operating points.

\subsection{Quality over Generation}
\begin{figure}[!t]
 \centering
 \includegraphics[width=\textwidth,trim=0 26bp 0 41bp,clip]{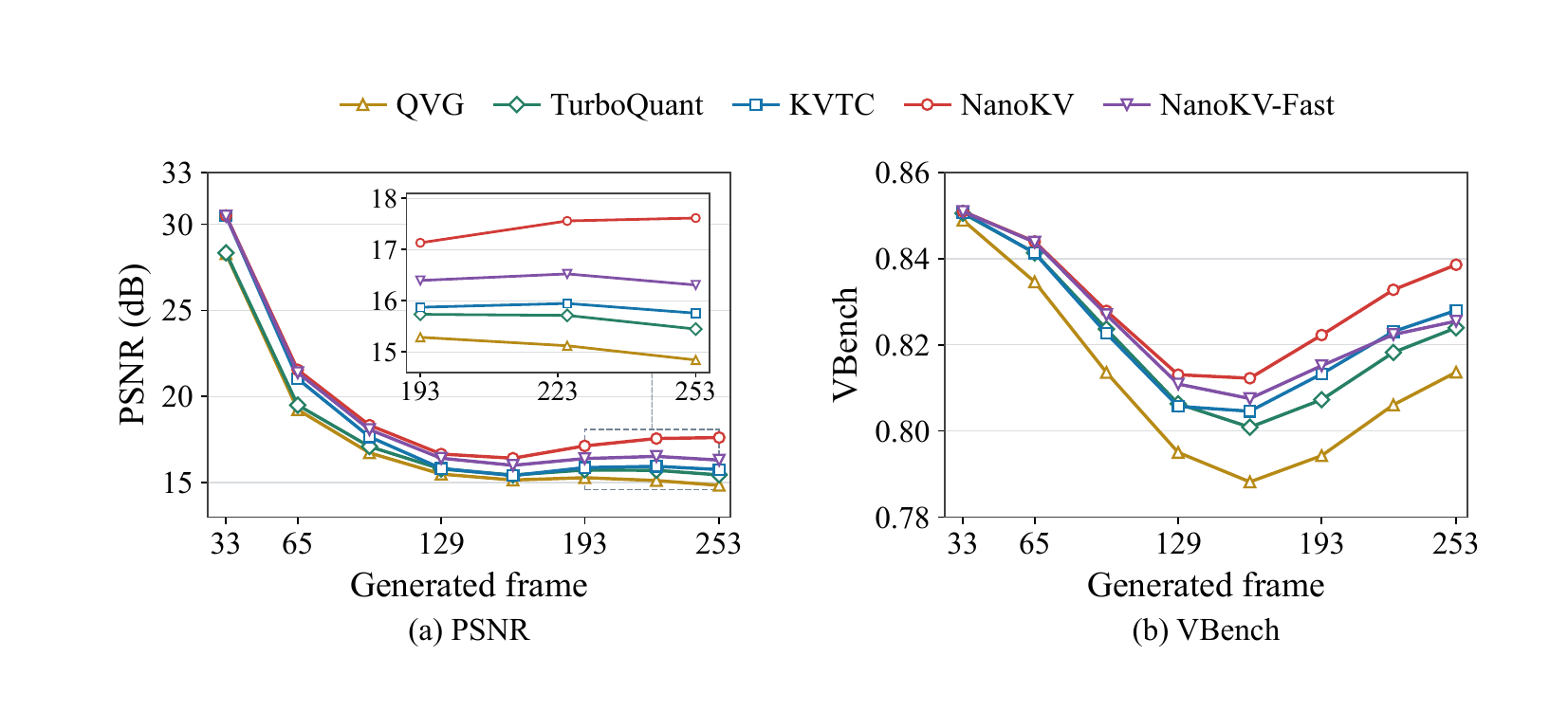}
 \caption{\textbf{Preserving quality over continuous generation.} On HY-WorldPlay at nominal $32\times$, NanoKV maintains the highest (a) reference PSNR and (b) VBench among evaluated compression methods throughout generation; NanoKV-Fast's gap grows later in the rollout.}
 \label{fig:temporal-vbench5}
\end{figure}

Figure~\ref{fig:temporal-vbench5} examines whether quality gains persist throughout HY-WorldPlay rollouts. NanoKV maintains the highest reference PSNR and VBench among the evaluated compressed methods, extending its advantage beyond early frames.
NanoKV-Fast follows a similar trajectory, although its gap from NanoKV grows later in generation, suggesting accumulation of low-dimensional approximation errors.

\section{Conclusion}
\label{sec:conclusion}
We presented NanoKV, a generation-aware framework for high-ratio KV compression with a frozen video backbone. Joint head--channel residual coding provides a compact representation, while role-specific fitting preserves block outputs for Keys and reconstructed content through quantization for Values. NanoKV-Fast reuses the same coefficients for mixed-dimensional attention, extending storage compression to computation reduction. Across three autoregressive video backbones, NanoKV improves the quality--storage trade-off. On HY-WorldPlay, it achieves 26.09$\times$ effective KV compression; NanoKV-Fast retains 20.53$\times$ compression and 19.04 dB PSNR with approximately 1.15$\times$ end-to-end speedup over KVTC.

\paragraph{Limitations.}
\label{sec:limitations}
Codec fitting and low-dimensional attention adaptation remain backbone-specific. Resident operators reduce effective compression, especially for short histories, while NanoKV-Fast's low-dimensional approximation may accumulate errors and further degrade quality over long horizons.

\clearpage
\subsection*{Ethics Statement}
NanoKV reduces the storage and computation costs of autoregressive video generation. While this can improve accessibility, lower deployment costs may also facilitate misuse, including deceptive media and non-consensual content. Our compression framework does not address the biases or safety limitations of the underlying video models, and fidelity to an uncompressed rollout is not a measure of content safety. Responsible use should respect source model and dataset licenses, protect personal data, and clearly identify synthetic content where appropriate. Deployment requires content safeguards beyond the compression techniques studied here.

\subsection*{AI Use Statement}
Generative AI tools assisted implementation and debugging, experiment automation, analysis and method discussions, literature organization, drafting portions of the manuscript, and language polishing. The authors led the research, designed the methods, established the experimental protocols, drew the conclusions, and independently validated the code, results, references, and all claims in this paper.

\subsection*{Reproducibility Statement}
Section~\ref{sec:method} describes the codec and mixed-dimensional attention. Section~\ref{sec:experiments} defines the evaluation metrics; Appendices~\ref{app:reproducibility}--\ref{app:additional-evaluation} provide the cache boundary, training objectives, positional transforms, and storage accounting. We will release NanoKV's implementation, fitted codec parameters for each evaluated video model, NanoKV-Fast configurations, and evaluation scripts.

\bibliography{references}
\bibliographystyle{nanokv_preprint}

\clearpage
\appendix
\suppressfloats[t]
\section{Common Cache and Baseline Protocol}
\label{app:reproducibility}

\label{app:cache-protocol}
We compress and persist keys before positional encoding, then apply the original positional transforms layer by layer when reading the cache. Original, QVG, TurboQuant, KVTC, and our method use this common cache implementation, including the same KV-cache lifecycle optimizations within each model. Main-table KV compression therefore measures the additional storage reduction from each codec.

\paragraph{Pre-position cache.}
Keeping keys before positional encoding lets positional branches share historical features, as in pre-RoPE key quantization~\citep{kvquant2024}. On HY-WorldPlay with 32 historical latent frames, this reduces persistent BF16 visual KV from 41.92 to 20.96 GiB. Reads still apply the original positions, frequencies, and camera transforms.

\paragraph{KV-cache lifecycle management.}
HY-WorldPlay releases expired storage before cache rebuilds; Matrix-Game 3.5 manages history by segment to avoid full-cache concatenation; LingBot-World 2.0 uses an exact fused RoPE read to avoid a full-history complex intermediate. All preserve the original history selection, current-chunk KV, and text KV.

\paragraph{Memory and output consistency.}
Table~\ref{tab:cache-boundary-audit} reports these changes. Completed comparisons have zero decoded-pixel error.

HY-WorldPlay's native 32-frame-history configuration is OOM on the A100 80GB. The table's estimated 114 GiB peak assumes two additional 20.96 GiB cache copies overlap with the measured 71.75 GiB pre-position peak; lifecycle optimization lowers the measured peak to 55.92 GiB.

\begin{table}[tbp]
\caption{Cache implementation: output consistency, runtime, and peak allocated memory.}
\label{tab:cache-boundary-audit}
\centering
\setlength{\tabcolsep}{4pt}
\begin{tabular}{llrrr}
\toprule
Model & Change & Pixel error & $\Delta$Time & Peak (GiB)\\
\midrule
HY-WorldPlay & post- $\to$ pre-RoPE & -- & -- & \begin{tabular}[c]{@{}c@{}}OOM ($\approx114$, est.)\\$\to$ 71.75\end{tabular}\\
 & KV lifecycle optimization & 0 & $-1.60\%$ & 71.75 $\to$ 55.92\\
LingBot-World 2.0 & post- $\to$ pre-RoPE & 0 & $-2.07\%$ & 59.33 $\to$ 59.33\\
Matrix-Game 3.5 & KV lifecycle optimization & 0 & $+1.69\%$ & 62.56 $\to$ 41.38\\
\bottomrule
\end{tabular}
\end{table}

\begin{figure}[tbp]
\centering
\includegraphics[width=\textwidth]{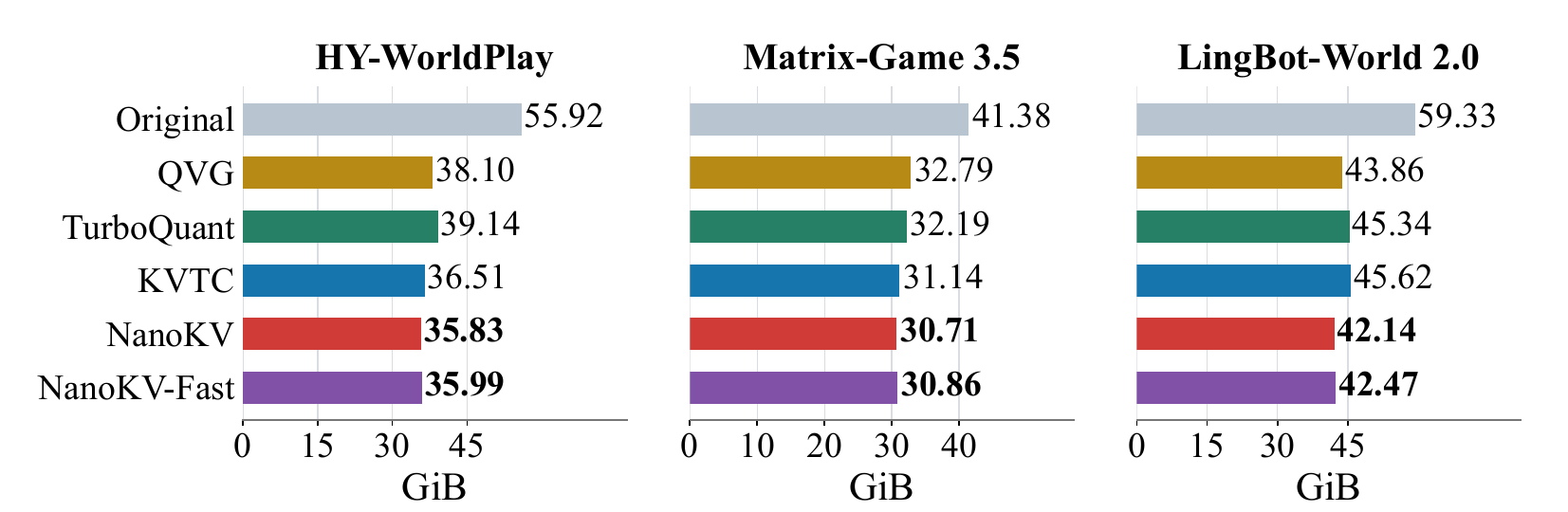}
\caption{PyTorch peak allocated GPU memory (GiB) for all methods on the fixed timing input.}
\label{fig:peak-memory-baselines}
\end{figure}

Figure~\ref{fig:peak-memory-baselines} compares end-to-end peaks for all methods under the common cache protocol.

\paragraph{Runtime reference at 20-frame history.}
At 20 historical latent frames, HY-WorldPlay's native and optimized implementations produce identical decoded pixels. Runtime changes from 278.88 to 298.54 seconds ($+7.05\%$), while peak allocated memory falls from 77.89 to 47.87 GiB. This comparison uses the shorter history because native caching is OOM at the formal 32-frame setting.

\paragraph{Baseline configurations.}
Table~\ref{tab:baseline-configuration} lists the evaluated implementations and resident numerical formats.
Our quantitative comparisons keep the pretrained backbone fixed, so reference fidelity measures deviations from the same generator. This setting permits codec fitting, but not backbone retraining. VideoMLA~\citep{videomla2026} changes the attention architecture and trains the adapted generator through teacher forcing and distillation. We therefore discuss it as an architectural alternative in Figure~\ref{fig:compression-designs}, rather than include it in the frozen-backbone comparison; assessing adapted generators would require a separate training-budget and generation-quality protocol.
\begin{table}[tbp]
\caption{Evaluated baseline implementations. Protection here denotes additional uncompressed historical tokens, not the model's common history window.}
\label{tab:baseline-configuration}
\centering
\small
\begin{tabular}{@{}>{\raggedright\arraybackslash}p{.14\linewidth}>{\raggedright\arraybackslash}p{.25\linewidth}>{\raggedright\arraybackslash}p{.25\linewidth}>{\raggedright\arraybackslash}p{.25\linewidth}@{}}
\toprule
Method & Implementation / coding & Grouping and protection & Resident precision \\
\midrule
QVG & Official Quant-VideoGen backend; INT2 residuals & Block size 64; 256 centers each for K/V; two K-means iterations; one residual stage; no extra raw history & BF16 centers and scales (A100 adaptation) \\
TurboQuant & Our implementation of the paper's algorithms; K: 2-bit + 1-bit residual; V: 3-bit & Per-head 128D transforms; no extra raw history & FP32 transforms/codebooks; BF16 norms \\
KVTC & Our video port; PCA and DP mixed-bit allocation; packed INT2/INT4/FP8, with zero-bit omissions & Consecutive layer groups; four sink and 128 recent tokens & BF16 PCA bases, scales and shifts; FP32 means \\
\bottomrule
\end{tabular}
\end{table}
KVTC uses separate K/V bases. Its group size and maximum rank are 8/4096 and 8/2048 for HY-WorldPlay at $16\times$ and $32\times$, 8/5120 for LingBot-World 2.0, and 6/2304 for Matrix-Game 3.5. The online video implementations use bit packing without entropy coding. All methods retain the common history policy in Section~\ref{app:cache-protocol}. Resident operators are included in KV CR.

\section{Compression Representation and Codec Training}
\label{app:compression-details}

\subsection{Controlled comparison of compression dimensions}
\label{app:axis-comparison}
Joint head--channel PCA retains the most KV variance in Figure~\ref{fig:motivation}(a). To isolate the compression axis, all four variants use $8\times$ dimensional compression with PCA, without quantizing the compressed coefficients. We fit on 48 trajectories from 24 training images and validate on four trajectories from two held-out images, covering left--right and forward--backward actions.

All variants use pre-RoPE K/V from layers 0, 10, 20, 30, 40, and 53 and chunks 5, 8, 11, and 15. Capture uses seed 1, four denoising steps, 253 output frames, 20 retrieved history positions, 12 recent context positions, with identical INT8 staging and dequantization for all variants. This shared capture step is separate from the compression operations being compared. Each projection subtracts the training mean and reconstructs with its PCA basis transpose.

For equal coefficient counts, all methods use a $30\times52$ crop of the $30\times53$ spatial grid. Spatial PCA reduces height and width by $2\times4$; spatiotemporal PCA reduces time, height, and width by $2\times2\times2$. Per-head PCA maps 128 channels to 16; joint PCA maps 2048 head--channel features to 256. The axis-wise methods use separate bases, and no variant quantizes PCA coefficients or uses subsequent adaptation.

Held-out KV variance retained is measured relative to the training mean:
\begin{equation}
 \mathrm{KV\ variance\ retained}=100\left(1-\frac{\|X-\widehat X\|_F^2}
 {\|X-\mu_{\mathrm{train}}\|_F^2}\right).
\end{equation}
Each layer uses 192 training and 16 validation snapshots. We accumulate validation errors before scoring each layer, then average six layers separately for K and V. Actual storage, including fitted operators, is evaluated separately.

\subsection{Training data and codec fitting}
\label{app:codec-objectives}
The codec is fitted separately for each layer while the video backbone remains frozen. Fitting uses 48 trajectories from 24 training images; within each model, our codec and KVTC use the same source trajectories and sampled data. HY-WorldPlay fitting uses a 20-position history, while its formal video evaluation uses 32 historical latent frames.

\begin{table}[tbp]
\caption{Codec and Key-alignment settings. Rank is per layer and per K/V branch; centers are fitted separately for K and V.}
\label{tab:codec-configuration}
\centering
\small
\begin{tabular}{lrrrrr}
\toprule
Model & Budget & Rank & Centers & K learning rate & Inner updates \\
\midrule
HY-WorldPlay & $16\times$ & 512 & 64 & $5\cdot10^{-6}$ & 5 \\
HY-WorldPlay & $32\times$ & 256 & 64 & $5\cdot10^{-6}$ & 5 \\
LingBot-World 2.0 & $32\times$ & 640 & 64 & $2.5\cdot10^{-6}$ & 1 \\
Matrix-Game 3.5 & $32\times$ & 384 & 64 & Not used & Not used \\
\bottomrule
\end{tabular}
\end{table}
All codecs use two K-means iterations and INT4 coefficients. Value fitting uses AdamW with learning rate $10^{-4}$, zero weight decay, batch size 512, and 12,000 updates per layer. Key alignment also uses AdamW, with the learning rates and inner updates in Table~\ref{tab:codec-configuration}; it supervises chunks 1--4 for HY-WorldPlay and chunks 2, 4, 6, 8 for LingBot-World 2.0. Each group visits all 48 training trajectories.

\paragraph{Quantized coefficients.}
Token-wise symmetric INT4 quantization stores the compressed coefficients with a BF16 scale. For coefficients $z_i$, the scale, integer code, and reconstructed coefficients are
\begin{equation}
 s_i=\frac{\max(\|z_i\|_\infty,\epsilon)}{7},\qquad
 q_i=\operatorname{clip}\!\left(\operatorname{round}(z_i/s_i),-7,7\right),
 \qquad \hat z_i=s_iq_i .
\end{equation}
Here $\epsilon>0$ is the numerical floor used by the implementation. Codes are packed into four bits; all-zero coefficient vectors produce zero codes.

\paragraph{Layer-wise error propagation.}
\label{app:layer-error-propagation}
Let $F_\ell(H;M)$ denote layer $\ell$ with input $H$ and historical KV $M$, with other conditioning held fixed. The reference and compressed paths satisfy $H_{\ell+1}=F_\ell(H_\ell;M_\ell)$ and $\hat H_{\ell+1}=F_\ell(\hat H_\ell;\hat M_\ell)$. For $\delta_\ell=\hat H_\ell-H_\ell$, define the local KV-induced error at the compressed path's input as
\[
 \epsilon_\ell=F_\ell(\hat H_\ell;\hat M_\ell)-F_\ell(\hat H_\ell;M_\ell).
\]
Adding and subtracting $F_\ell(\hat H_\ell;M_\ell)$ gives
\[
 \begin{aligned}
 \delta_{\ell+1}
 &=F_\ell(\hat H_\ell;M_\ell)-F_\ell(H_\ell;M_\ell)+\epsilon_\ell\\
 &\approx J_\ell\delta_\ell+\epsilon_\ell,\qquad
 J_\ell=\left.\frac{\partial F_\ell(H;M_\ell)}{\partial H}\right|_{H=H_\ell}.
 \end{aligned}
\]
The approximation expands the reference layer around $H_\ell$. It separates propagation of the existing input error from the new error induced by compressed history within one denoising forward pass.

\paragraph{Block-output error diagnostic.}
Figure~\ref{fig:motivation}(b) compares HY-WorldPlay at nominal $32\times$ compression with and without Key block-output alignment on eight videos from two images and four actions. For each block and chunk, we average $\|\hat Y-Y\|_2/\max(\|Y\|_2,10^{-12})$ over matched output samples from the eight videos and four denoising steps. Lines and shading summarize the resulting 16 chunk values by their mean and population standard deviation. Averaging all 54 blocks and 16 chunks gives errors of 0.2667 without alignment and 0.2260 with alignment; their relative reduction is 15.3\%.

\paragraph{Key block-output alignment.}
The losses in Section~\ref{sec:key-learning} are averaged over each layer group, with a separate Key decoder updated for every layer. HY-WorldPlay uses six groups of nine layers; LingBot-World 2.0 uses five groups of eight, trained from shallow to deep. Encoders, codes, and backbone remain fixed.

The implemented losses use relative reconstruction error $r(X,\hat X)=\|\hat X-X\|_F/\max(\|X\|_F,10^{-8})$. Let $c_t$ denote mean token-wise cosine distance and $c_f$ cosine distance after flattening. The per-layer terms are
\begin{equation}
\begin{aligned}
\text{HY-WorldPlay:}\quad
 \mathcal L_{\rm out}&=r(Y,\hat Y)+0.1c_t(Y,\hat Y),\\
 \mathcal L_{K,\rm rec}&=0.2r(K,\hat K)+0.02c_f(K,\hat K);\\
\text{LingBot-World 2.0:}\quad
 \mathcal L_{\rm out}&=\frac{\|\hat Y-Y\|_F^2}{\max(\|Y-H_s\|_F^2,10^{-8})}+0.02c_f(Y,\hat Y),\\
 \mathcal L_{K,\rm rec}&=0.2r(K,\hat K).
\end{aligned}
\label{eq:key-loss-details}
\end{equation}
Here $H_s$ is the compressed path's input to the current block; the denominator measures the reference block's update relative to that input. Output losses use 256 sampled tokens and 256 channels; reconstruction uses up to 65,536 entries per cache tensor. Losses are averaged over supervised layers and cache segments.

\paragraph{Quantization-aware Value fitting.}
The Value encoder and decoder learn to reconstruct residuals after quantization, starting from random orthogonal maps rather than PCA. For the original residual $R^V$ and its reconstruction $\hat R^V$, we use
\begin{equation}
 \mathcal L_V=
 \frac{\|R^V-\hat R^V\|_1}{\|R^V\|_1}
 +0.2\frac{\|R^V-\hat R^V\|_2^2}{\|R^V\|_2^2}
 +1-\cos(R^V,\hat R^V).
 \label{eq:value-loss}
\end{equation}
Here $\cos(R^V,\hat R^V)$ is cosine similarity along the feature dimension, averaged over tokens. The forward pass includes quantization; the backward pass approximates the rounding gradient by the identity so that both linear maps can be updated. The backbone and key branch remain fixed, and loss denominators are numerically stabilized.

\paragraph{Fitting cost.}
KV fitting takes approximately 1--2 hours; serial Key alignment finishes within one day per model on one A100. Data collection and low-dimensional adaptation are separate stages.

\paragraph{Role of shared centers.}
\label{app:center-covariance}
Shared centers store the common component of tokens in each cluster, leaving the low-dimensional coefficients to represent differences between tokens rather than repeatedly encoding shared content. The first two rows of Table~\ref{tab:component-ablation} show the effect of adding this representation to joint PCA with INT4 quantization.

\paragraph{Component effects across backbones.}
NanoKV improves reference fidelity while using less KV storage on both HY-WorldPlay and LingBot-World 2.0 (Table~\ref{tab:main-results}), supporting generation-aware compression beyond a single backbone. The HY-WorldPlay ablation further isolates the gains from role-specific K/V fitting (Table~\ref{tab:component-ablation}). Differences in cache organization and attention structure motivate backbone-specific fitting settings (Table~\ref{tab:codec-configuration}); the magnitude of each component's benefit can vary while retaining the same principle of preserving how historical KV affects generation.

\section{NanoKV-Fast}
\label{app:lowd-details}
NanoKV-Fast demonstrates that NanoKV's stored coefficients can also support lower-dimensional attention without retraining the backbone. The common mechanism is compressed-coefficient decoding, mixed-dimensional computation, and LSE merging; dimensions and protection policies are backbone-specific operating choices. We do not propose a universal configuration or a fully automatic end-to-end configuration-search algorithm. Below we describe offline map fitting and configuration selection, followed by the shared computation.

\subsection{Fitting low-dimensional maps and selecting configurations}
\label{app:learning-protocol}
NanoKV-Fast constructs mixed-dimensional attention from the same compressed coefficients. Its deployment configuration specifies the dimension of each head in each layer and which historical tokens retain full-dimensional computation. We first fit candidate low-dimensional maps, form an initial dimension allocation from local output errors, and then refine protection using generated-video quality. Map fitting and configuration selection are separate stages; the backbone and codec remain frozen.

\paragraph{From local error to video quality.}
All three backbones follow the same configuration-selection protocol. Low-dimensional maps target full-dimensional attention on the compressed cache, and local output errors provide candidate head-dimension allocations. We then use generated-video quality on development images disjoint from the evaluation set to select the dimension allocation and history-protection policy. The selected configuration is fixed across all evaluation samples. Changing protection reuses the fitted maps without retraining; the backbone and codec remain frozen.

\paragraph{Model-specific dimensions and protection.}
HY-WorldPlay uses per-head dimensions of 32, 64, 96, and 128, with full-dimensional fallback for sensitive layers and protection of recent history and periodic anchors.

LingBot-World 2.0 also uses 32, 64, 96, and 128 dimensions, with protection of initial chunks, sink tokens, and recent frames. Its initial allocation scores candidate dimensions by output-projection-weighted squared error, including the mean compensation in Section~\ref{app:latent-attention-details}, and uses error reduction per added channel.

Matrix-Game 3.5 uses 64 and 128 dimensions. Its initial allocation ranks head--layer pairs by attention-output error on held-out training calls. The selected configuration has 432 low-dimensional and 288 full-dimensional pairs and protects the five most recent cached latent frames, without additional whole-layer fallback, sink protection, or periodic anchors.

These model-specific candidates are selected under the same independent development protocol above. Tables~\ref{tab:deployed-dimensions} and~\ref{tab:protection-configuration} give the complete dimension allocations and protection rules. Protected history remains compressed in storage and is reconstructed to full dimension only when read.

\begin{table}[tbp]
\caption{Head--layer counts at each deployed attention dimension.}
\label{tab:deployed-dimensions}
\centering
\begin{tabular}{llrrrr}
\toprule
Model & \shortstack{Nominal\\coefficient CR} & $d_h=32$ & $64$ & $96$ & $128$\\
\midrule
HY-WorldPlay & 16$\times$ & 448 & 208 & 32 & 176 \\
HY-WorldPlay & 32$\times$ & 440 & 192 & 24 & 208 \\
LingBot-World 2.0 & 32$\times$ & 926 & 159 & 304 & 211 \\
Matrix-Game 3.5 & 32$\times$ & 0 & 432 & 0 & 288 \\
\bottomrule
\end{tabular}
\end{table}

\begin{table}[tbp]
\caption{Full-dimensional routing in the deployed NanoKV-Fast configurations. Layer indices are zero-based; frame counts refer to latent frames.}
\label{tab:protection-configuration}
\centering
\small
\begin{tabular}{p{.24\linewidth}p{.68\linewidth}}
\toprule
Model & History routed to full dimension \\
\midrule
HY-WorldPlay & Four recent frames; one anchor per 16 older frames, or per eight on layers 15, 17, 19, 20, 22, 23, 26, 30, 31, 36, 46, 48. Full-layer fallback: 27--35, 40, 45 at $16\times$; additionally 39 and 47 at $32\times$. \\
LingBot-World 2.0 & First three chunks; one sink frame and two recent frames thereafter. Layer 11 uses a full-dimensional prefix for all eight chunks. \\
Matrix-Game 3.5 & Five recent cached latent frames; sensitive head--layer pairs retain 128 dimensions. No additional sink or periodic anchors use full-dimensional reads. \\
\bottomrule
\end{tabular}
\end{table}

\paragraph{Map-fitting settings.}
Map fitting follows each backbone's attention and positional structure, with settings given below. Matrix-Game 3.5 fits its codec on the 48 trajectories in Section~\ref{app:codec-objectives}, but fits its low-dimensional maps separately on four training trajectories.

\paragraph{HY-WorldPlay.}
Low-dimensional maps are trained against full-dimensional attention outputs while replaying the corresponding frozen blocks. The analytically initialized configuration in Table~\ref{tab:lowd-adaptation} keeps the deployed head dimensions, selected RoPE pairs, and full-dimensional routing. The low-dimensional Key decoder takes the corresponding codec-decoder columns with an identity Query selection; Value uses a per-head truncated SVD, and the PRoPE maps use group-preserving SVD. Attention adaptation then optimizes these maps with normalized attention-output, centered-logit, KL, and frozen-block-output losses, weighted $1$, $0.20$, $0.05$, and $0.25$. It uses AdamW with an initial learning rate of $2\times10^{-4}$ and selects the adapted maps by validation block-output error.

\paragraph{Matrix-Game 3.5.}
Low-dimensional attention maps are fitted against the frozen NanoKV codec with 64 shared centers, rank-384 coefficients, and Value adaptation. Four training trajectories provide 16 captured attention calls per layer. The last call of each trajectory is held out, leaving 12 fitting and four validation calls. Each of the 30 layers uses 800 Adam updates at a learning rate of $10^{-4}$, with gradient norms clipped to one. The loss combines normalized attention-output squared error, KL divergence from full-dimensional to low-dimensional attention, and squared log-sum-exp error with weights $1$, $0.1$, and $0.02$. We select weights by held-out loss and fold the learned KV projections into the codec decoders to decode low-dimensional KV directly from compressed coefficients.

At deployment, current tokens and mosaic context use full-dimensional attention. Both branches retain the original $1/\sqrt{128}$ score scale and merge through LSE after output-dimension restoration.

\paragraph{LingBot-World 2.0.}
We fit the maps from attention statistics. Key and Query use complete RoPE pairs. Each pair contributes an attention-logit feature. Its covariance is weighted by attention probability and the mass $m$ assigned to history; a mean-logit term has weight $m(1-m)+0.05$. Greedy least squares selects $d_h/2$ pairs using a ridge coefficient of $10^{-4}$ times the mean covariance diagonal. Selected pairs define the Key map; fitted Query weights are clipped to $[0,2]$.

For Value and output, we take the leading $d_h$ eigenvectors of the historical-mass-weighted covariance of mean-centered reference attention outputs, then use their transpose for restoration.

\paragraph{Effect of attention adaptation.}
On the 20-video HY-WorldPlay panel used for the $32\times$ quality--compute trade-off, attention adaptation raises PSNR by 1.38 dB over the analytical initialization of the low-dimensional maps (Table~\ref{tab:lowd-adaptation}). The codec, head dimensions, full-dimensional routing, and attention arithmetic are identical in both rows. The same compressed cache decoded to full dimension reaches 18.96 dB on this panel; adaptation brings the low-dimensional path to 18.59 dB.

\begin{table}[tbp]
\caption{NanoKV-Fast low-dimensional attention adaptation on the HY-WorldPlay $32\times$ panel. Higher PSNR/SSIM and lower LPIPS are better.}
\label{tab:lowd-adaptation}
\centering
\setlength{\tabcolsep}{8pt}
\begin{tabular}{lrrr}
\toprule
Low-dimensional maps & PSNR $\uparrow$ & SSIM $\uparrow$ & LPIPS $\downarrow$\\
\midrule
Analytical initialization & 17.21 & 0.5463 & 0.3451\\
+ Attention adaptation & \textbf{18.59} & \textbf{0.5920} & \textbf{0.2837}\\
\bottomrule
\end{tabular}
\end{table}

\subsection{Implementing mixed-dimensional attention}
\label{app:latent-attention-details}
Low-dimensional attention approximates the full-dimensional computation from the same compressed coefficients. This section details the factorization, positional structure, and branch merging of NanoKV-Fast introduced in Section~\ref{sec:latent-attention}.

\paragraph{Attention-aware factorization.}
The useful low-dimensional directions depend on queries and historical features together. Omitting positional and affine terms, the full decoder reconstructs head $h$ as $\hat K_h=Z_KW_{\mathrm{dec},h}^K$ and $\hat V_h=Z_VW_{\mathrm{dec},h}^V$, where $W_{\mathrm{dec},h}^K,W_{\mathrm{dec},h}^V\in\mathbb R^{r\times D}$ are head-specific column blocks of $W_{\mathrm{dec}}$. The unscaled full-dimensional key product is
\[
 Q_h\hat K_h^\top=Q_h(W_{\mathrm{dec},h}^K)^\top Z_K^\top.
\]
For this factorization, we absorb the score calibration $\alpha_h$ into the Query map $A_h$. The low-dimensional path instead uses $A_h\in\mathbb R^{D\times d_h}$ to reduce the query and $B_h^K\in\mathbb R^{r\times d_h}$ to decode the coefficients, giving the unscaled product $Q_hA_h(B_h^K)^\top Z_K^\top$. Matching these products for arbitrary $Q_h$ and $Z_K$ would require
\begin{equation}
 A_h(B_h^K)^\top=(W_{\mathrm{dec},h}^K)^\top.
 \label{eq:exact-key-condition}
\end{equation}
On the value side, let $W_h$ denote the attention weights. The full output is $W_hZ_VW_{\mathrm{dec},h}^V$, whereas the low-dimensional path computes $W_hZ_VB_h^VL_h$: $B_h^V$ decodes values to $d_h$ dimensions and $L_h\in\mathbb R^{d_h\times D}$ restores the aggregated output. Exact preservation would therefore require $B_h^VL_h=W_{\mathrm{dec},h}^V$.

Both matrix products have rank at most $d_h$, so the low-dimensional path generally cannot reproduce the full decoders exactly. To reduce this approximation error, we fit the maps using queries and compressed histories observed during generation rather than selecting directions solely from the decoder spectrum. Section~\ref{app:learning-protocol} gives the fitting procedure for each model.

\paragraph{Center offsets and value-mean compensation.}
The offsets in the main method are the low-dimensional forms of the center and fixed mean. For token $i$,
\begin{equation}
 \bar C_{h,i}^K=(c_{a_i,h}^K+\mu_h^K)P_h^K,\qquad
 \bar C_{h,i}^V=(c_{a_i,h}^V+\mu_h^V)P_h^V,
 \label{eq:lowdim-center-offset}
\end{equation}
Here $P_h^K,P_h^V\in\mathbb R^{D\times d_h}$ are head-specific offset maps, while $c_{a_i,h}^{K/V}$ and $\mu_h^{K/V}$ are the $D$-dimensional head slices of the stored center and fixed mean. Thus, the stored full-dimensional centers are mapped directly rather than relearned; K and V retain separate center assignments. LingBot-World 2.0 additionally restores the part of the historical value mean lost during low-dimensional mapping:
\begin{equation}
 O_h^{\mathrm{low}}=\bar O_hL_h+m_h-\bar m_hL_h,
\end{equation}
where $m_h$ is the mean of the full-dimensional reconstructed history assigned to the low-dimensional branch and $\bar m_h$ its low-dimensional mean. This correction is applied to that branch before LSE merging. Coefficient sums and center counts provide these means without reconstructing the full history.

\paragraph{Positional structure.}
Complete positional groups preserve the original transforms within the retained coordinates. RoPE uses complete channel pairs with their original frequencies. HY-WorldPlay's PRoPE uses four-coordinate camera groups; Matrix-Game 3.5's paired camera layout requires aligned eight-dimensional blocks and applies the original output camera transform before restoring the head dimension. LingBot-World 2.0 preserves RoPE pairs. All three models retain their token positions and original transform order.

\paragraph{Joint softmax across branches.}
Log-sum-exp statistics combine full- and low-dimensional branches into one attention distribution. For each query row and branch $s\in\{n,l\}$, let $S_s$ be its logits, $\Lambda_s=\log\sum_j\exp S_{s,j}$ its normalizer, and $O_s=\sum_j e^{S_{s,j}-\Lambda_s}V^{\mathrm{eff}}_{s,j}$ its branch-normalized output. Here $V^{\mathrm{eff}}_{s,j}$ includes any low-dimensional output lift and mean correction, so both branches have $D$-dimensional values. Splitting the joint softmax denominator gives
\begin{equation}
 \Lambda=\operatorname{logaddexp}(\Lambda_n,\Lambda_l),\qquad
 O=e^{\Lambda_n-\Lambda}O_n+e^{\Lambda_l-\Lambda}O_l .
 \label{eq:lse-merge}
\end{equation}
The merge is exact for the logits produced by the two branches; attention adaptation controls the approximation in the low-dimensional branch.

\section{Additional Evaluation and Visual Evidence}
\label{app:additional-evaluation}

\subsection{Evaluation and deployment accounting}
\label{app:evaluation-protocol}
Table~\ref{tab:model-protocol} supplements the evaluation setup with model dimensions and video lengths.

\begin{table}[tbp]
\caption{Model dimensions and evaluation video lengths.}
\label{tab:model-protocol}
\centering
\begin{tabular}{lrrrr}
\toprule
Model & Layers & Heads & Head dimension & Video frames\\
\midrule
HY-WorldPlay & 54 & 16 & 128 & 253\\
LingBot-World 2.0 & 40 & 40 & 128 & 125\\
Matrix-Game 3.5 & 30 & 24 & 128 & 169\\
\bottomrule
\end{tabular}
\end{table}

\paragraph{Image splits and inference settings.}
The source images come from the VBench-I2V image suite. A category-stratified, image-level split of a fixed 200-image pool uses selection seed 20260809, without model-quality feedback. The 24 codec-fitting images and 100 evaluation images are disjoint; Fast configuration development is described separately in Appendix~\ref{app:learning-protocol}. The four evaluation actions are left--right rotation, right--left rotation, forward--backward motion, and left--right translation.

HY-WorldPlay generates $848\times480$ videos with four denoising steps and seed 1; LingBot-World 2.0 generates $832\times464$ videos with four steps and seed 42; Matrix-Game 3.5 generates $1280\times704$ videos with three steps and seed 3407. Within each backbone, methods use matched image, prompt, action, and seed. HY-WorldPlay retains 32 historical latent frames. LingBot uses the native settings \texttt{local\_attn\_size=18} and \texttt{sink\_size=6}; Matrix-Game uses seven three-latent-frame context chunks with its boundary-anchor and mosaic-context policy. Image identifiers and action specifications are recorded in the fitting and evaluation manifests.

\paragraph{Quality and operating points.}
The HY-WorldPlay component analysis and Figure~\ref{fig:compute-tradeoff} use the same 20 videos from five images and four actions. The quality--compute analysis varies the protection policy to select, for each historical portion, either its fitted low-dimensional representation or full-dimensional reconstruction. This changes the history range and its assigned attention dimensions in Section~\ref{sec:experiments}, without refitting the low-dimensional maps or changing their output dimensions. Quality metrics follow the main evaluation and exclude the common first frame.

\paragraph{Persistent KV storage.}
\label{app:deployment}
For NanoKV, the nominal coefficient compression ratio is $16HD/(4r)=4HD/r$, comparing BF16 features with INT4 coefficients and excluding auxiliary storage. Effective KV CR additionally counts compressed history metadata and resident operators. History includes packed INT4 coefficients, BF16 scales, center indices, and centers. Exported codec weights use row-wise INT8 with BF16 scales; resident BF16 low-dimensional maps and enabled encoder or decoder copies are also counted. Matrix-Game 3.5 routing buffers are also included. Shared tensors, such as means shared with the codec, are counted once.

\begin{figure}[tbp]
  \centering
  \includegraphics[width=\textwidth]{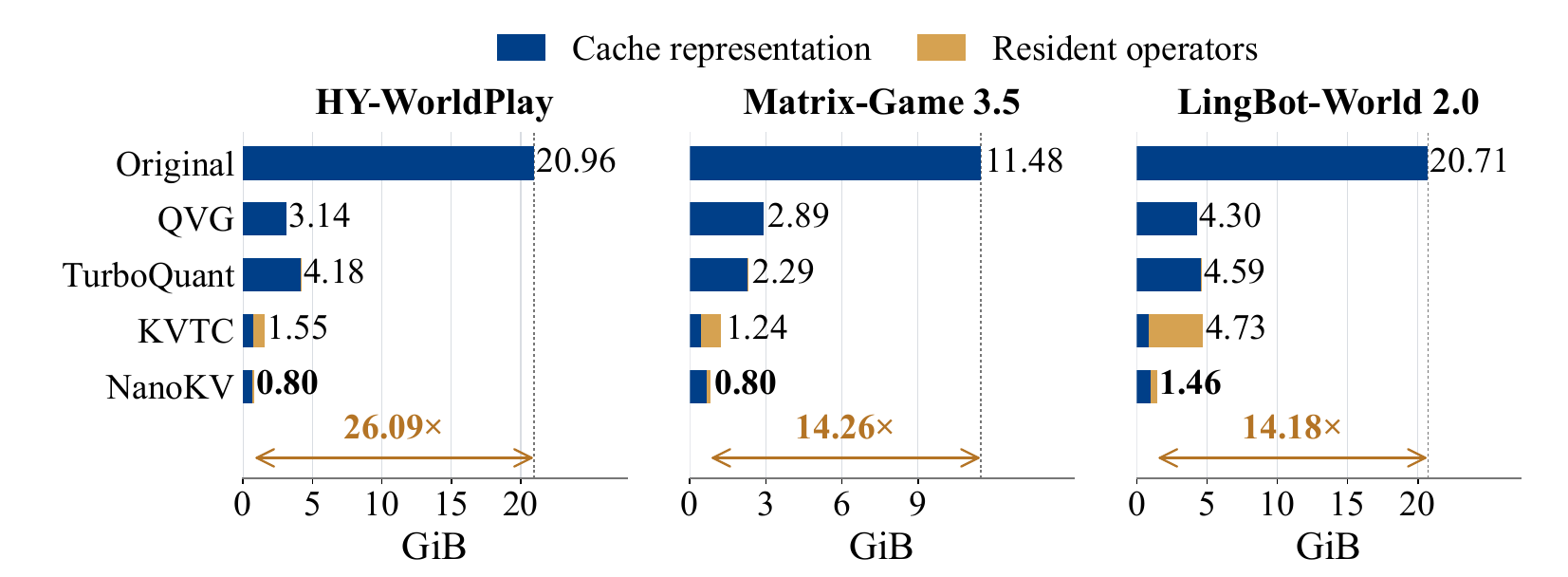}
  \caption{Persistent KV storage (GiB) for full-dimensional attention configurations, split into cached data and resident operators.}
  \label{fig:system-storage}
\end{figure}

\paragraph{Runtime and peak memory.}
\label{app:runtime-memory}
Runtime covers online generation, including VAE decoding but excluding model loading; Matrix-Game 3.5 timings also include video writing. For the fixed timing input, latencies are the medians of three valid runs; the main table reports Original's latency and each method's speedup relative to Original. Figures~\ref{fig:teaser} and~\ref{fig:peak-memory-baselines} use \texttt{torch.cuda.max\_memory\_allocated()} over this inference interval, counting live tensors and PyTorch-managed workspaces.

Table~\ref{tab:peak-memory-crosscheck} reports NVML peaks over the monitored run, including model loading, allocator reservations, and CUDA allocations outside PyTorch. Allocated memory measures active tensor demand; NVML reflects the broader capacity needed for deployment. The largest gap is 7.29 GiB for HY-WorldPlay KVTC. Because the intervals differ and peaks may occur at different times, their difference is not a direct measure of allocator overhead.

\begin{table}[tbp]
\caption{NVML-measured peak GPU memory (GiB), complementing the PyTorch allocated peaks in Figure~\ref{fig:peak-memory-baselines}.}
\label{tab:peak-memory-crosscheck}
\centering
\setlength{\tabcolsep}{8pt}
\renewcommand{\arraystretch}{1.04}
\begin{tabular}{lrrr}
\toprule
Method & HY-WorldPlay & Matrix-Game 3.5 & LingBot-World 2.0 \\
\midrule
Original & 58.70 & 42.41 & 60.40 \\
Quant VideoGen & 40.90 & 34.00 & 46.39 \\
TurboQuant & 42.27 & 33.34 & 47.16 \\
KVTC (32$\times$) & 43.80 & 32.21 & 47.94 \\
NanoKV (32$\times$) & 38.56 & 31.96 & 44.56 \\
NanoKV-Fast (32$\times$) & 39.46 & 32.13 & 44.89 \\
\bottomrule
\end{tabular}
\end{table}

\paragraph{Reconstruction cost.}
\label{app:reconstruction-cost}
In Matrix-Game 3.5, KVTC reuses decoded coefficients across consecutive layers in a group, but reconstructs each layer with its own $2304\!\times\!3072$ basis slice. NanoKV uses a $384\!\times\!3072$ decoder. Repeated layer-wise reconstruction therefore involves larger matrix products for KVTC during the segmented rollout.

\paragraph{Low-dimensional efficiency.}
Table~\ref{tab:compute-breakdown} distinguishes savings within historical attention from savings over all visual attention. Historical savings use full-dimensional attention over all retained history as the denominator, including history routed to full dimension but excluding current tokens and cache recomputation. Multiplying this saving by the historical share gives the overall visual-attention MAC reduction, not the end-to-end latency reduction. Tables~\ref{tab:deployed-dimensions} and~\ref{tab:protection-configuration} specify each model's dimensions and protection policy.

Figure~\ref{fig:compute-tradeoff}(b) measures savings in historical-attention arithmetic, not wall-clock time. End-to-end gains are limited by the fraction of runtime spent in this attention: other computation, including feed-forward layers and VAE decoding, remains unchanged. Decoding, projection, routing, and kernel overhead further reduce the realized gain. Thus, arithmetic savings need not yield proportional latency reductions, and can coexist with a small net slowdown, as on LingBot-World 2.0.

The Matrix-Game 3.5 implementation fuses positional transforms and selected-row dequantization, and caches routing metadata to reduce layout copies and repeated synchronization. Complete-video checks confirm that these implementation optimizations preserve the frozen low-dimensional output pixel for pixel.

\begin{table}[tbp]
\caption{Denoising visual-attention arithmetic of NanoKV-Fast relative to NanoKV. The overall saving is the product of the historical share and the saving within history.}
\label{tab:compute-breakdown}
\centering
\setlength{\tabcolsep}{5pt}
\begin{tabular}{llrrr}
\toprule
Model & Budget & \shortstack{Historical\\share} & \shortstack{History MACs\\saved$\uparrow$} & \shortstack{All visual-attention\\MACs saved$\uparrow$}\\
\midrule
HY-WorldPlay & $16\times$ & 85.19\% & 39.36\% & 33.53\%\\
HY-WorldPlay & $32\times$ & 85.19\% & 37.93\% & 32.31\%\\
LingBot-World 2.0 & $32\times$ & 71.43\% & 34.32\% & 24.51\%\\
Matrix-Game 3.5 & $32\times$ & 72.00\% & 20.83\% & 15.00\%\\
\bottomrule
\end{tabular}
\end{table}

\subsection{Quality over generation}
\label{app:extended-analysis}
Figures~\ref{fig:temporal-vbench-extra} and~\ref{fig:temporal-psnr} supplement the main results with HY-WorldPlay at $16\times$ and LingBot-World 2.0 at $32\times$. We track quality over consecutive intervals of up to 32 generated frames, excluding the common first frame, and place each point at its interval's final frame. The shorter LingBot sequence uses half-overlapping intervals. VBench and reference-based PSNR use the same intervals to describe perceptual quality and divergence from the matched Original rollout.

\IfFileExists{figures/temporal_vbench5_appendix.pdf}{%
\paragraph{VBench over generation.}
VBench uses the same five dimensions and averaging rule as the main evaluation, displayed on a 0--100 scale.
\begin{figure*}[!htbp]
 \centering
 \includegraphics[width=\textwidth]{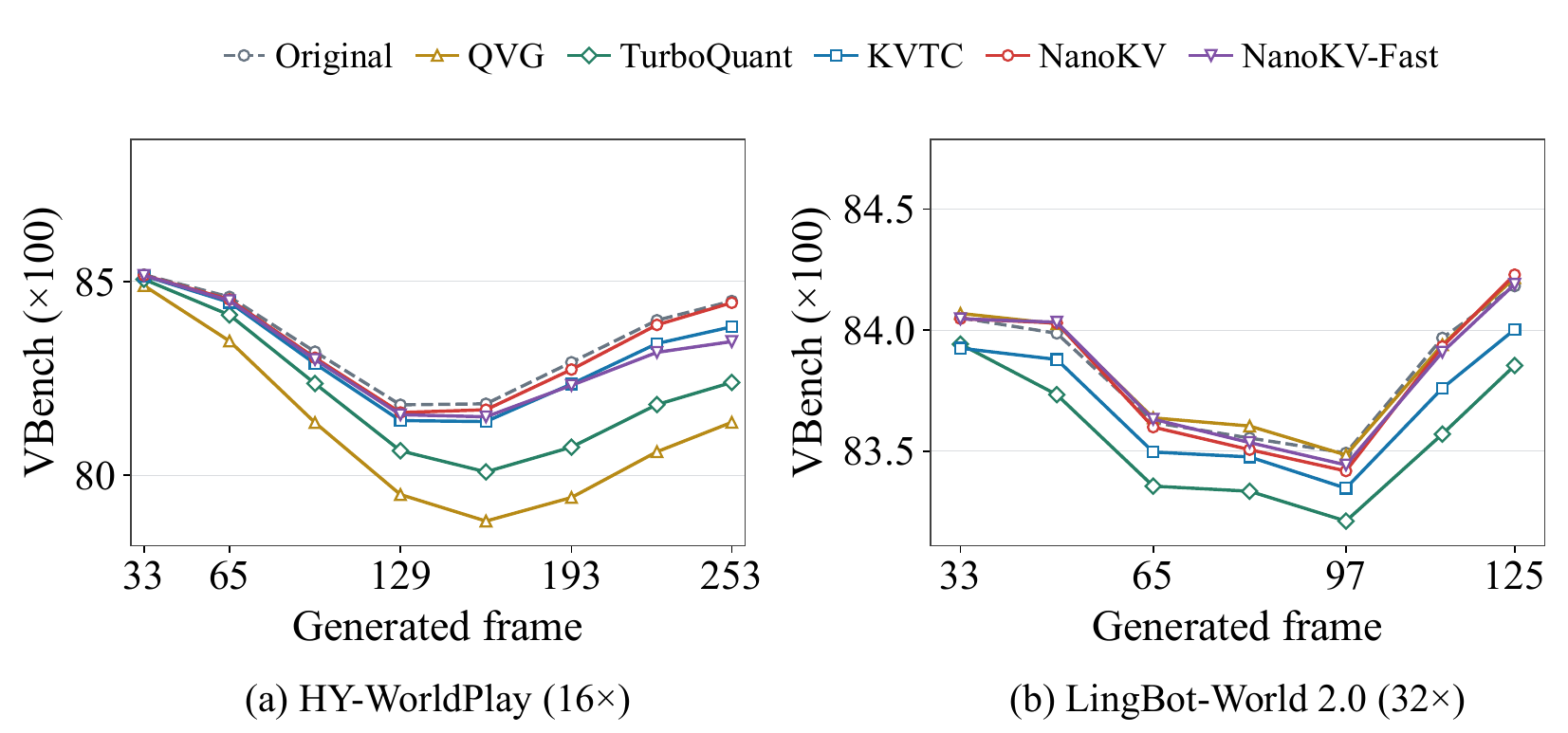}
 \caption{VBench over generation for (a) HY-WorldPlay at $16\times$ and (b) LingBot-World 2.0 at $32\times$.}
 \label{fig:temporal-vbench-extra}
\end{figure*}
}{}

\IfFileExists{figures/temporal_psnr_appendix.pdf}{%
\paragraph{Reference fidelity over time.}
We compare the first, middle, and final generated frames of each local interval with matched Original frames. PSNR is computed separately for each sampled frame, averaged within intervals, and then averaged across videos; all methods use the same frame positions.
\begin{figure*}[!htbp]
 \centering
 \includegraphics[width=\textwidth]{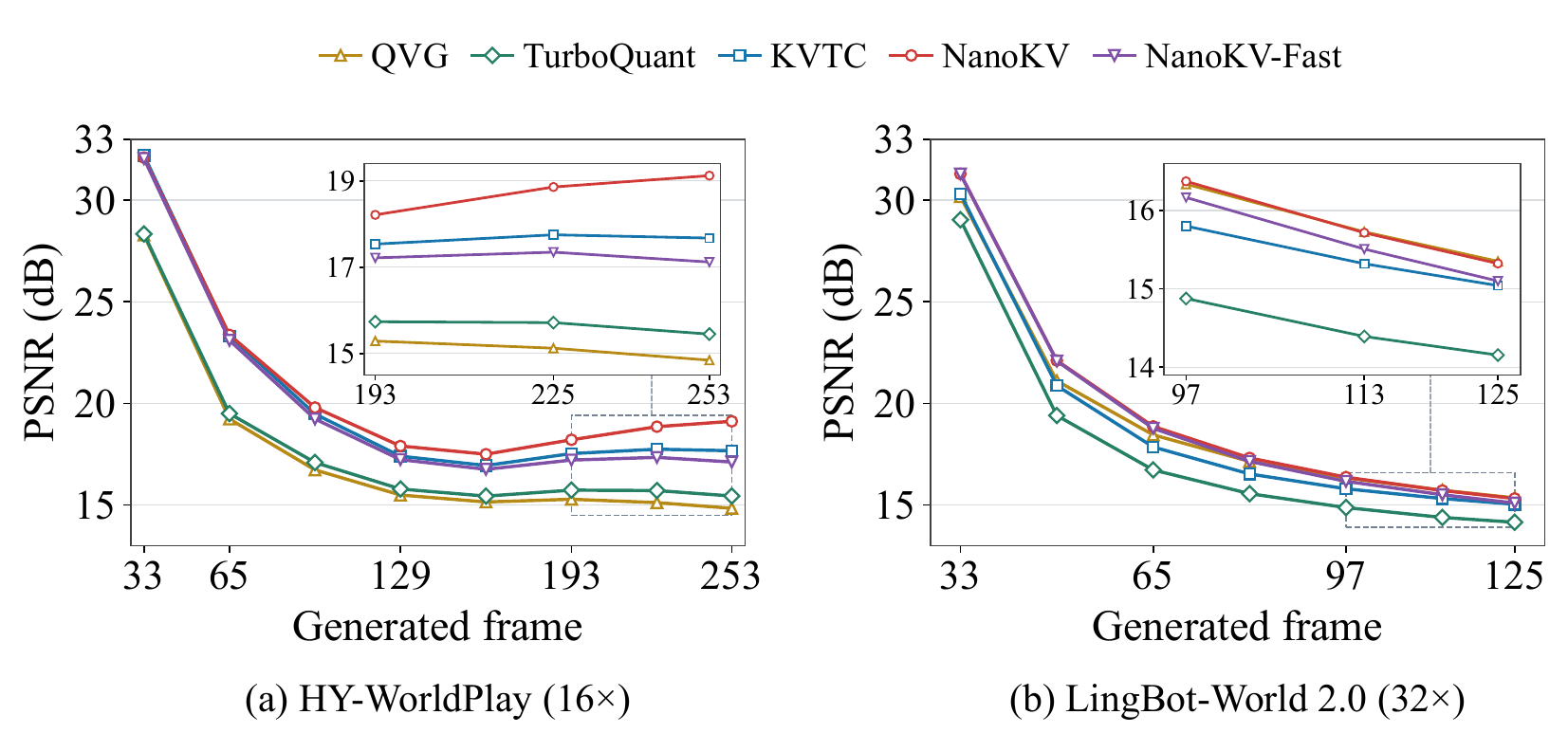}
 \caption{Reference fidelity over generation for (a) HY-WorldPlay at $16\times$ and (b) LingBot-World 2.0 at $32\times$.}
 \label{fig:temporal-psnr}
\end{figure*}
}{}

\clearpage
\subsection{Additional qualitative examples}
\label{app:video-comparisons}
The additional examples show scene structure and appearance across four camera-action sequences for each model. We use twelve distinct input images and match the first, middle, and final frames across methods. The $32\times$ results use the main-table configurations; supplementary $16\times$ results are included where labeled.

\subsubsection{HY-WorldPlay}
\begin{figure*}[!htbp]
 \centering
 \includegraphics[width=\textwidth]{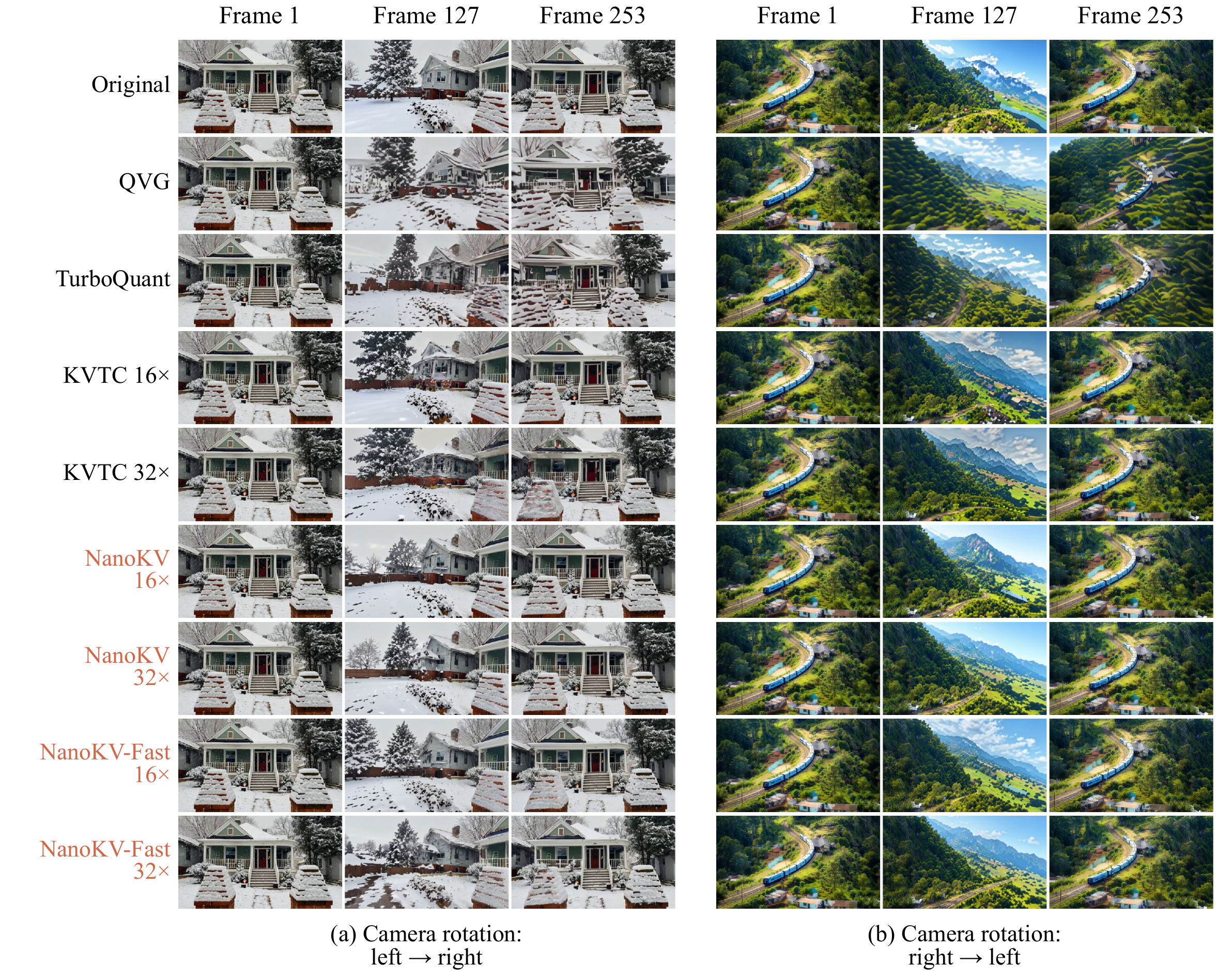}
 \caption{HY-WorldPlay examples under four camera actions, showing matched first, middle, and final frames. Rows compare Original, quantization baselines, KVTC, NanoKV, and NanoKV-Fast; both nominal compression budgets are included where labeled. Panels (c)--(d) continue on the next page.}
 \label{fig:video-worldplay-extra}
\end{figure*}

\clearpage
\begin{figure*}[tbp]
 \centering
 \includegraphics[width=\textwidth]{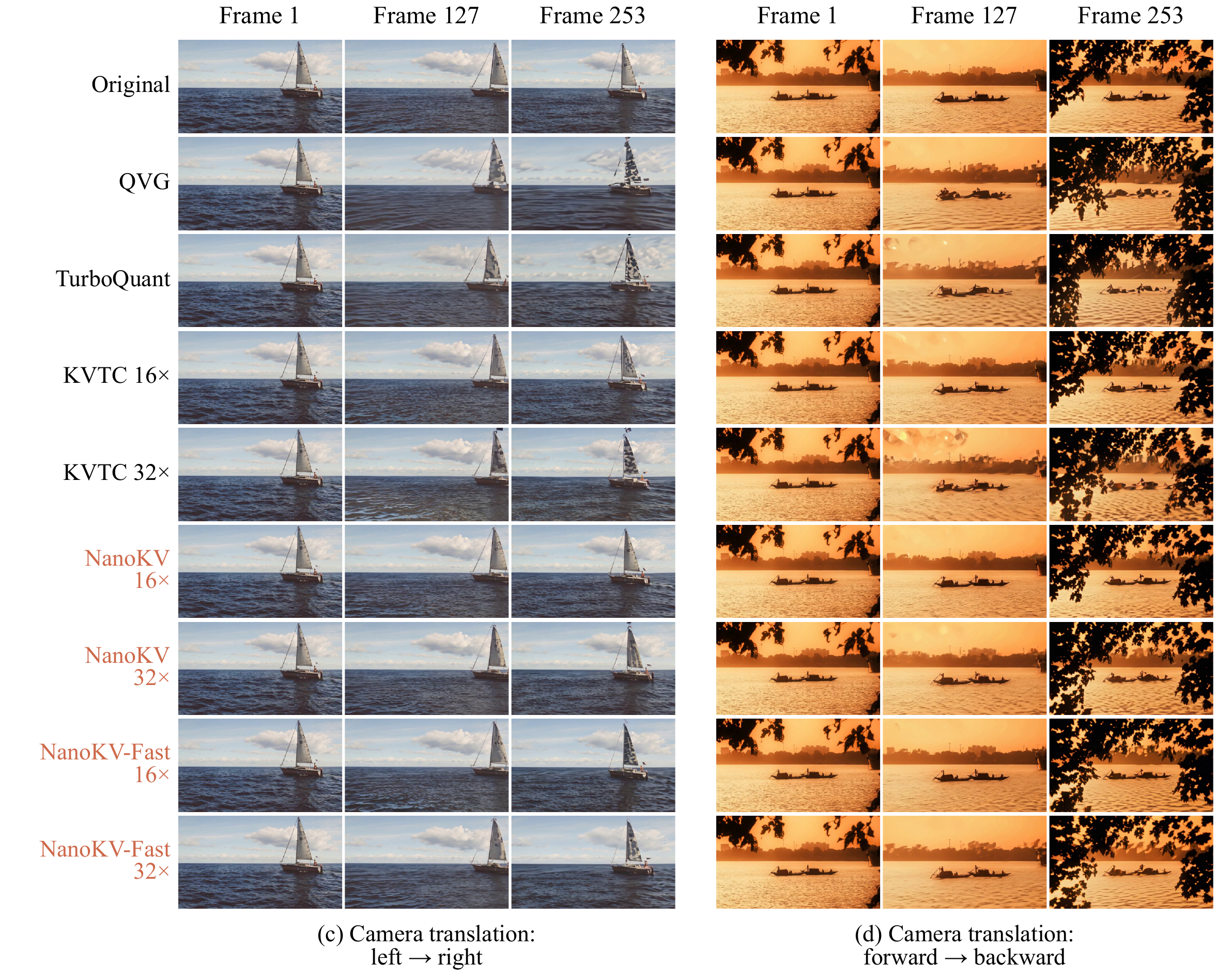}
 \par\medskip\textbf{Figure~\ref{fig:video-worldplay-extra} (continued).}
\end{figure*}

\clearpage
\subsubsection{LingBot-World 2.0}
\begin{figure*}[!htbp]
 \centering
 \includegraphics[width=\textwidth]{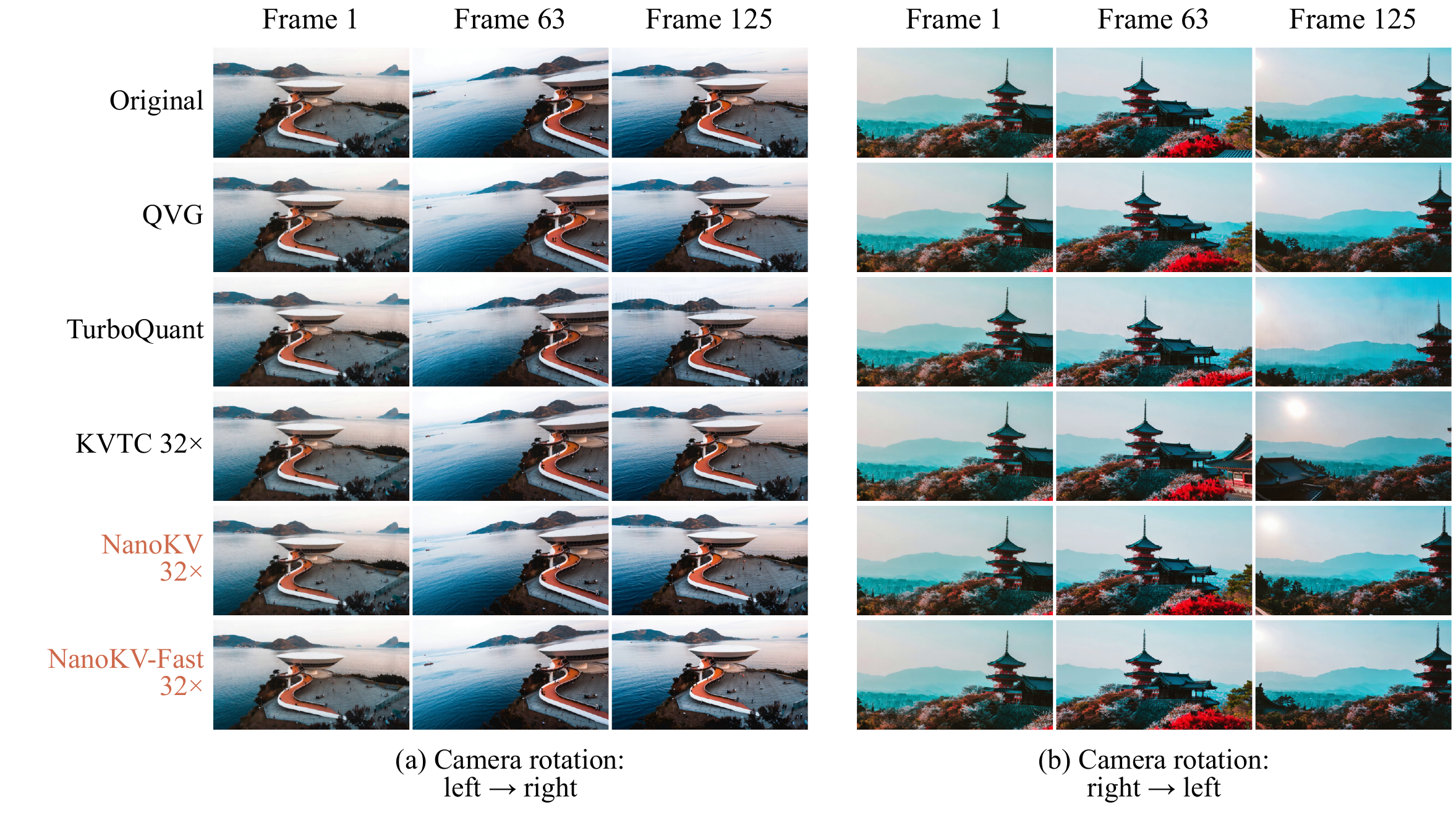}
 \par\medskip
 \includegraphics[width=\textwidth]{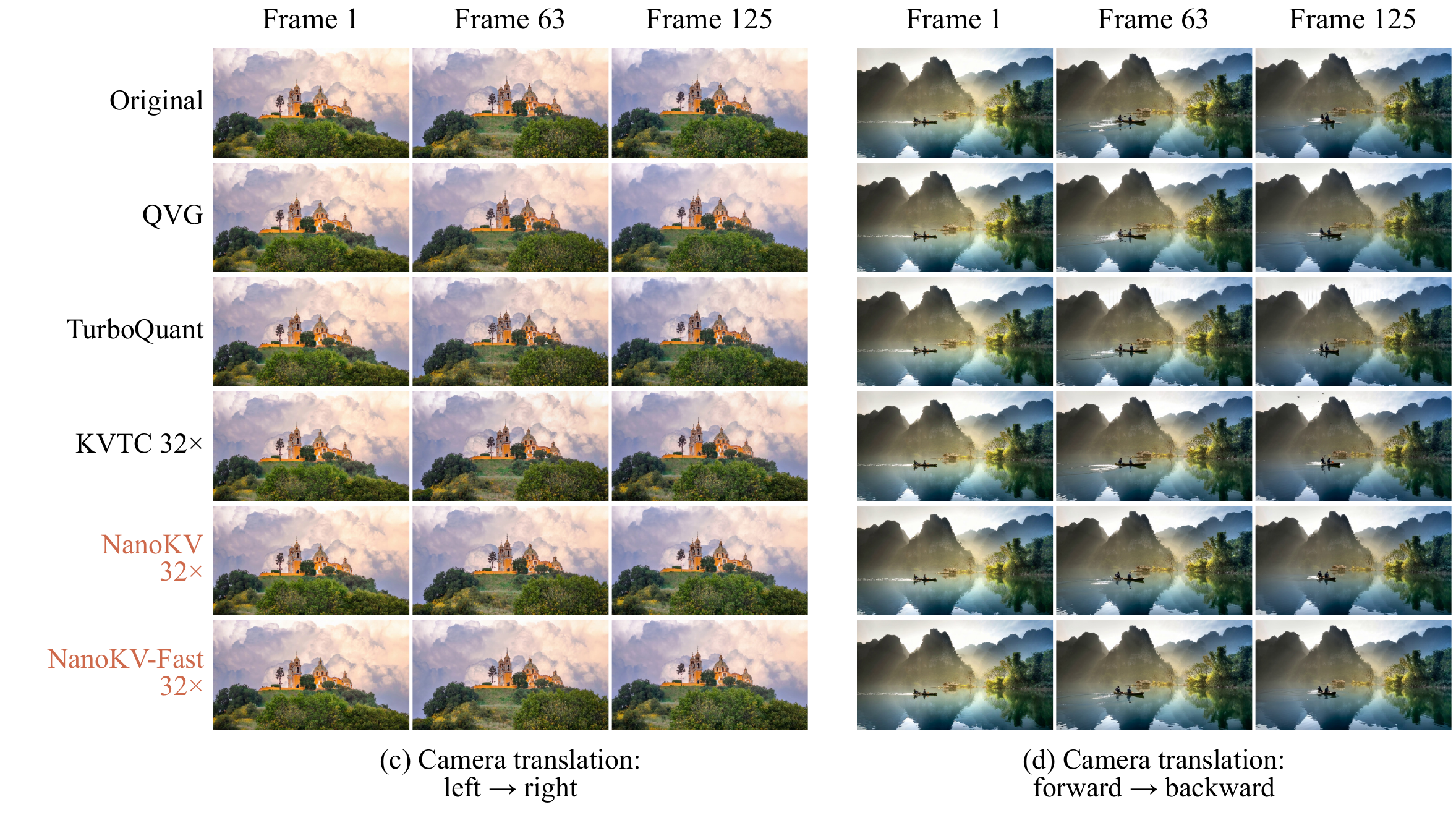}
 \caption{LingBot-World 2.0 examples under four camera actions. Matched first, middle, and final frames compare Original, QVG, TurboQuant, KVTC, NanoKV, and NanoKV-Fast. Full frames show scene consistency through camera motion and return.}
 \label{fig:video-lingbot-extra}
\end{figure*}

\clearpage
\subsubsection{Matrix-Game 3.5}
\begin{figure*}[!htbp]
 \centering
 \includegraphics[width=\textwidth]{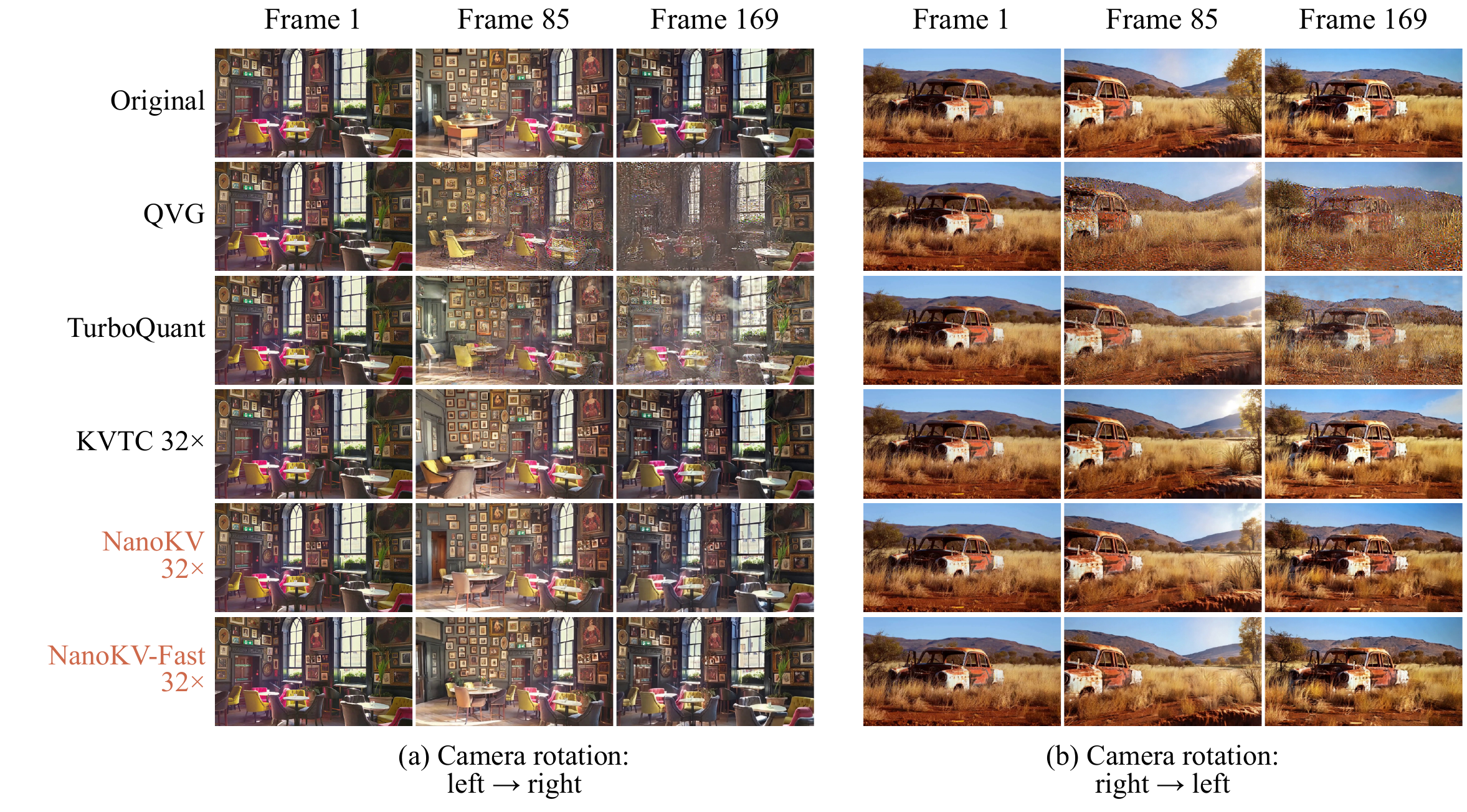}
 \par\medskip
 \includegraphics[width=\textwidth]{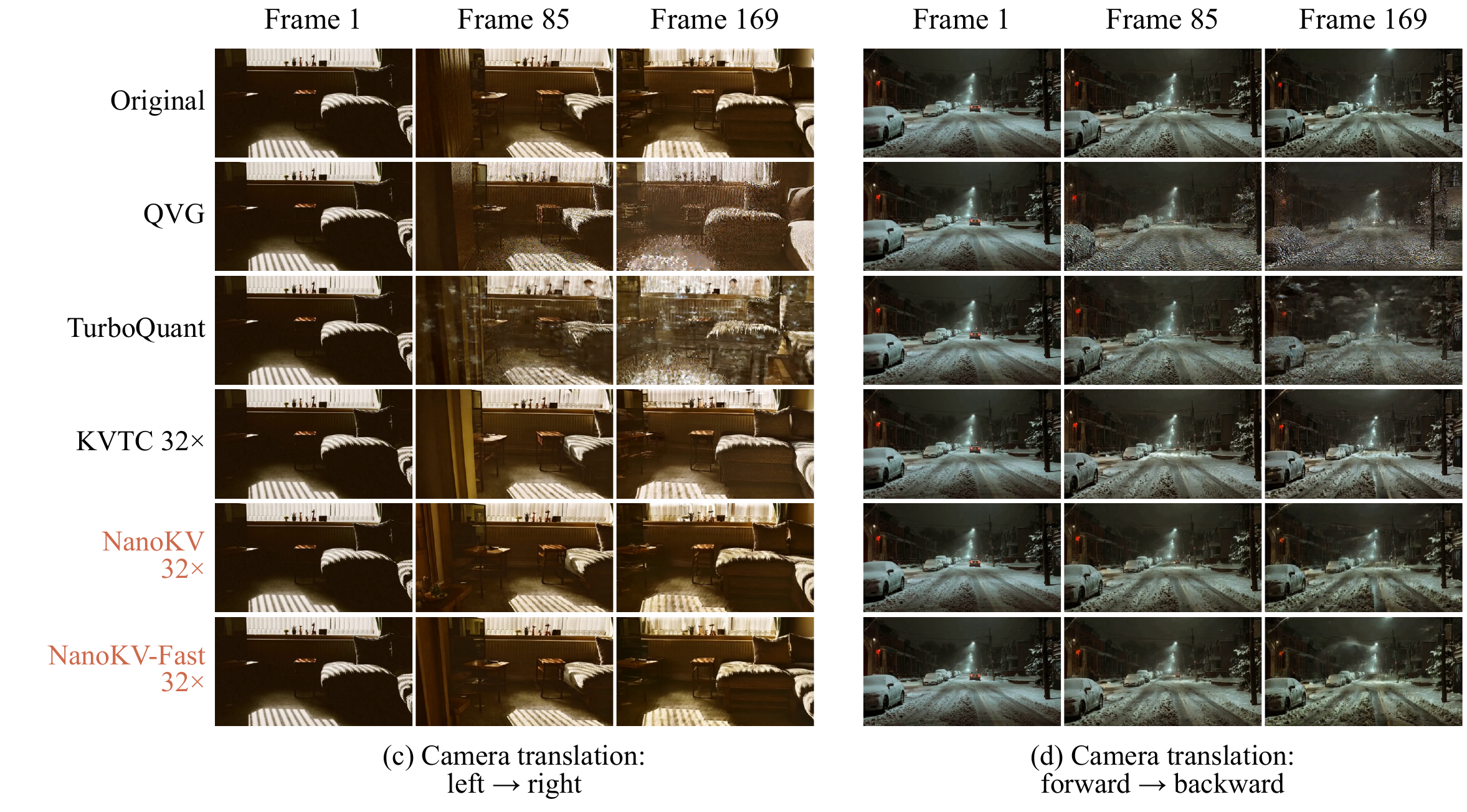}
 \caption{Matrix-Game 3.5 examples under four camera actions. Matched first, middle, and final frames compare Original, QVG, TurboQuant, KVTC, NanoKV, and NanoKV-Fast. All rows use the same input image and action sequence within each panel.}
 \label{fig:video-mg35-extra}
\end{figure*}

\end{document}